\documentclass[conference]{IEEEtran}
\IEEEoverridecommandlockouts

\usepackage{cite}
\usepackage{amsmath,amssymb,amsfonts}
\usepackage{textcomp}
\usepackage{xcolor}
\def\BibTeX{{\rm B\kern-.05em{\sc i\kern-.025em b}\kern-.08em
    T\kern-.1667em\lower.7ex\hbox{E}\kern-.125emX}}

\usepackage{subcaption}
\usepackage{url}
\usepackage[hidelinks]{hyperref}
\usepackage{booktabs}
\usepackage{placeins}

\usepackage{times}
\usepackage{soul}
\usepackage{url}
\usepackage[utf8]{inputenc}
\usepackage{graphicx}
\usepackage{amsmath}
\usepackage{amsthm}
\usepackage{booktabs}
\usepackage{makecell}
\usepackage{algorithm}
\usepackage[switch]{lineno}
\usepackage{mathtools}

\usepackage{latexsym}

\usepackage{amssymb}
\usepackage{tabularx}
\usepackage{multirow}
\usepackage{subcaption}
\usepackage{nicefrac}
\usepackage{enumitem}
\usepackage{algpseudocode}
\usepackage{xcolor}
\usepackage[most]{tcolorbox}
\usepackage{bbm}

\usepackage{pifont}

\newcommand{\RR}{\mathbb{R}}

\newcommand{\gdinoscore}{0.2}

\begin{document}

\makeatletter

\newcommand{\linebreakand}{%
  \end{@IEEEauthorhalign}
  \hfill\mbox{}\par
  \mbox{}\hfill\begin{@IEEEauthorhalign}
}

\title{Open-Vocabulary Models and Label Propagation\\ for Uncertainty-aware Pseudo-Labeling\\ of Object Detection Datasets
}

\author{
  \IEEEauthorblockN{1\textsuperscript{st} Jonathan Klees}
  \IEEEauthorblockA{\textit{Institute of Computer Science} \\
    \textit{Osnabrück University}\\
    jonathan.klees@uni-osnabrueck.de}
  \and
  \IEEEauthorblockN{2\textsuperscript{nd} Timo Heller}
  \IEEEauthorblockA{\textit{Department of Mathematics} \\
  \textit{University of Wuppertal} \\
    timo.heller@uni-wuppertal.de}
  \and
  \IEEEauthorblockN{3\textsuperscript{rd} Matthias Rottmann}
  \IEEEauthorblockA{\textit{Institute of Computer Science} \\
    \textit{Osnabrück University}\\
    matthias.rottmann@uni-osnabrueck.de}
}

\maketitle

\setlength{\parindent}{0pt}
\setlength{\parskip}{0.5em}

\begin{abstract}
Object detection models rely on large-scale annotated datasets that require substantial manual annotation effort.
The labeling costs further increase when annotation uncertainty is captured through soft labels, as multiple  annotations must be collected for each object.
We address this challenge by introducing a semi-supervised framework for the uncertainty-aware annotation of object detection datasets.
Our approach leverages open-vocabulary object detectors to generate region proposals and semantically meaningful object representations. 
Based on these representations, we construct an object-level similarity graph and propagate probabilistic label information.
Our framework assigns soft labels to previously unlabeled objects, quantifying both aleatoric and epistemic uncertainty.
We evaluate our method on four object detection benchmarks with soft-label annotations and demonstrate that high-quality annotations can be obtained with only a small fraction of manually labeled objects. 
Furthermore, we show that detectors trained on semi-automatically annotated data achieve performance comparable to models trained on fully manually annotated data.
We achieve state-of-the-art downstream detection performance for the task of semi-supervised object detection on the COCO dataset, with 40.5\% AP virtually matching the performance of the model trained on fully manually constructed data (40.9\%) at only 1\% of labeling costs.
These results highlight the potential of open-vocabulary models and graph-based label propagation for efficient, accurate and uncertainty-aware pseudo-labeling.
\end{abstract}

\section{Introduction}
\label{sec:intro}
% Annotation costs as a bottleneck
Modern object detectors rely on large-scale annotated datasets, whose creation requires substantial manual effort.
In contrast to image classification, object detection annotations must specify the location and category of every object, making dataset construction particularly time-consuming.
% Annotation uncertainty can only be captured with even higher annotation costs.
Furthermore, existing object detection benchmarks generally treat annotations as deterministic, despite the fact that ambiguity is inherent to many detection tasks due to occlusions, unclear object boundaries, blurring, and semantic uncertainty.
Recently, object detection datasets that capture semantic uncertainty were proposed~\cite{rechecked,benchmark}. Annotation uncertainty is captured through multiple independent annotations, often more than ten human evaluations per object. While such approaches provide valuable information about human uncertainty, their annotation costs may limit scalability.
% This motives SSL techniques to efficiently annotate OD datasets
This motivates the development of semi-automated methods that can efficiently propagate label information and capture semantic uncertainty.
While graph-based semi-supervised techniques proved to be effective in image classification~\cite{Zhou_2003,semi-supervised-learning,klees2026}, similar methods for uncertainty-aware annotation in object detection remain unexplored. 

\begin{figure}[t]
    \centering
    \includegraphics[width=\linewidth]{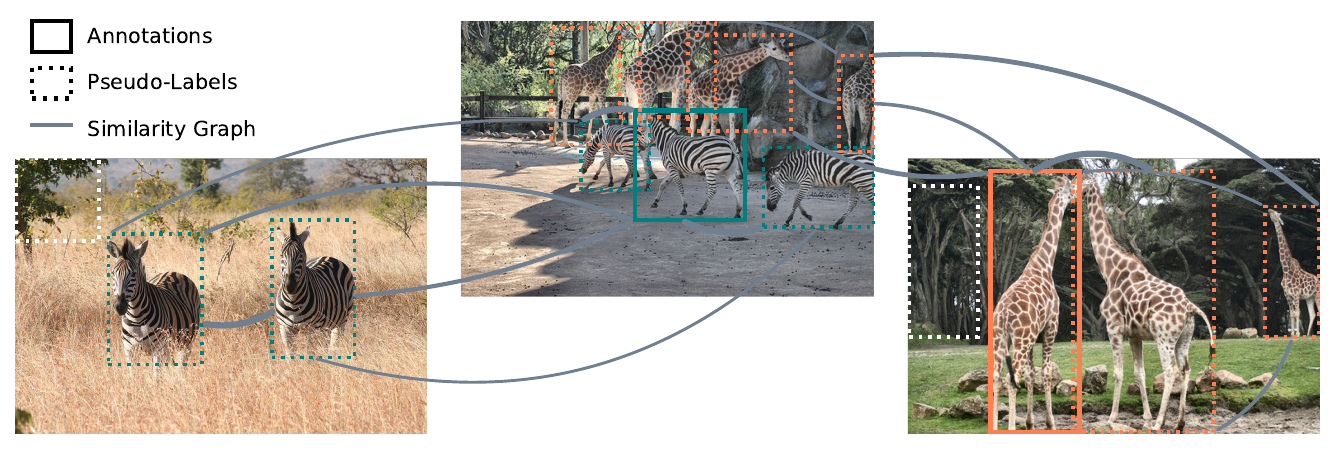}
    \caption{Illustration of label propagation for object detection datasets. Initial region proposals are generated with Grounding DINO~\cite{liu2023grounding} and form a similarity graph (grey lines) over which sparse label information (solid bounding boxes) is propagated. This results in pseudo-labels with
    soft label estimates based on which false positives are filtered effectively.
    }
    \label{fig:teaser}
\end{figure}

% What we do (high-level)
In this work, we develop graph-based semi-supervised learning in the domain of object detection by leveraging powerful open-vocabulary object detectors such as Grounding DINO~\cite{liu2023grounding} to generate initial region proposals and semantically meaningful object representations.
With these object representations, we construct a similarity graph over which label information is propagated (\autoref{fig:teaser}). To propagate label information, we rely on probabilistic label spreading~\cite{klees2026} which generalizes classical label spreading~\cite{Zhou_2003} to the domain of soft labels and accounts for aleatoric and epistemic uncertainty.
This way, our method propagates (potentially noisy) label information obtained for few objects to other localized but yet unlabeled objects. This enables the uncertainty-aware, semi-automated annotation of object detection datasets. As a result, all detected objects  are equipped with a soft label that reflects the individual class probabilities, including the background class.
Soft labels generalize conventional labels and reflect the aleatoric uncertainty associated with label assignments~\cite{nguyen_learning_2011}. They allow for uncertainty-aware training~\cite{Peng2014LearningOP} and thorough assessment of model uncertainty~\cite{benchmark}.
Creating soft labels manually requires the repeated evaluation of instances by multiple annotators so that datasets with soft labels are rare~\cite{klees2026,benchmark}. 

In the domain of object detection, \cite{benchmark} re-annotated the four object detection benchmarks COCO~\cite{COCO}, Pascal VOC~\cite{PascalVOC}, KITTI~\cite{kitti} and Cityscapes~\cite{Cityscapes} where they equipped objects with soft labels based on at least 11 human annotations. We evaluate our framework on these datasets and quantify achieved data quality in terms of average precision~\cite{COCO} and its soft label variant average similarity~\cite{benchmark}.
The results demonstrate that high-quality uncertainty-aware annotations can be obtained from only a small fraction of manually labeled objects, substantially reducing annotation effort while maintaining competitive data quality.
We can further reduce annotation effort by integrating active querying to steer the selection of objects to annotate next. We also show that our framework can effectively uncover label errors in existing datasets.
Notably, we observe that the quality of semi-automatically constructed data is sufficiently high for downstream tasks such as training closed-set object detectors. 
Consequently, our framework achieves state-of-the art downstream detection performance in semi-supervised object detection~\cite{sohn2020, soft_teacher} on the COCO dataset, with 40.5\% AP \emph{essentially matching} the fully supervised performance of 40.9\% at only 1\% of labeling costs.
%
% Contributions
\\
We summarize our main contributions as follows:
\begin{itemize}[left=0pt]
    % Method Contribution
    \item We introduce the first graph-based semi-supervised framework for uncertainty-aware annotation of object detection datasets. 
    It leverages open-vocabulary models and probabilistic label spreading to estimate soft labels of objects, capturing both aleatoric and epistemic uncertainty.
    % Performance
    \item We demonstrate the effectiveness of our method on four standard datasets. We obtain accurate annotations at a fraction of annotation effort where our method consistently outperforms baselines in terms of label quality.
    % Active Learning
    \item We integrate active querying to further reduce the annotation effort required to obtain accurate annotations.
    % Downstream Performance
    \item We show that training on semi-automatically constructed annotations yields comparable performance, compared to training on original annotations, setting a new state of the art in semi-supervised object detection on COCO.
\end{itemize}

% Method Figure
\begin{figure*}[t]
    \centering
    \includegraphics[
    width=0.95\linewidth,
    trim=0.05cm 1.25cm 0.1cm 0.75cm,
    clip]{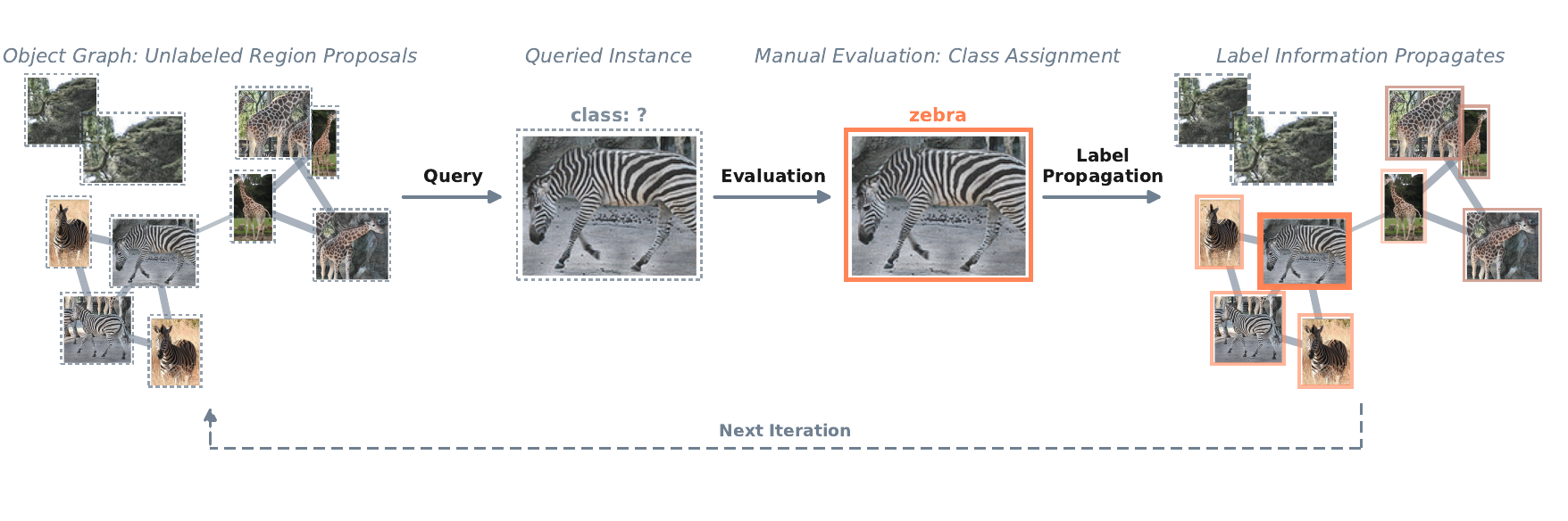}
    \caption{Method overview: Initial region proposals form a graph based on feature similarity and individual objects are queried for manual inspection. The queried object is assigned either one of the dataset's classes or the background class. This label information is then propagated over the similarity graph resulting in soft label estimates for each object proposal. The estimate is iteratively refined with each object that is inspected.}
    \label{fig:method}
\end{figure*}
\section{Related Work}

\paragraph{Graph-Based Semi-Supervised Learning.}
Semi-supervised learning (SSL) on graphs is an established research field with seminal contributions by~\cite{zhu2002,Zhou_2003}. 
Early work on \emph{label propagation}~\cite{zhu2002} combines labeled and unlabeled data in a graph whose edge weights are typically derived from a Gaussian kernel. Class information from labeled samples is then iteratively propagated to neighboring unlabeled nodes while preserving the original seed labels.
Building on this idea, \emph{label spreading}~\cite{Zhou_2003} relaxes the hard clamping of initial labels and performs a diffusion process over the graph, where a parameter $\alpha$ governs the balance between propagated and initial label information. 
Since their introduction, numerous extensions have been proposed, including formulations based on hypergraphs~\cite{nonlinear_label_spreading,higherorderlabelhomogeneityspreading}, methods designed for improved scalability and computational efficiency~\cite{fast_label_propagation,efficient_label_propagation}, and alternative strategies for graph construction. 
Comprehensive surveys of graph-based semi-supervised learning can be found in~\cite{van_engelen_survey_2020,semi-supervised-learning}. 
Beyond label propagation methods, annotation-efficient learning paradigms such as active learning and human-in-the-loop systems have been introduced to further reduce labeling costs in semi-automated annotation frameworks~\cite{Semi-Automated-Data-Labeling,cooperative_learning}.
While label propagation is well-studied in the domain of image classification, no approaches for object detection datasets exist. We propose the transfer of established graph-based SSL techniques to the domain of object detection by leveraging open-vocabulary models for initial region proposals and semantic object representations.

\paragraph{Label Propagation on an Object Level.}
Semi-automated annotation of objects is well-established for video data where objects are annotated based on only a few labeled  frames~\cite{Kao2016,Waldmann2022,chen2020,Chen2022,PseudoProp}.
Here, annotated objects are to be re-identified in other frames of the video to propagate initial label information. This is typically achieved by appropriate feature extraction and leveraging motion cues for temporal consistency.
The task is different from semi-automated annotation of object detection datasets since the data consists of similar scenes and labels are typically only propagated to the same objects across frames.

% Actual Label Propagation in OD
Wang et al.~\cite{Wang2017} conduct label propagation on an instance level. They construct a similarity graph over embeddings of segmented objects to propagate labels across instances using label spreading~\cite{Zhou_2003}. For this, the existence of segments is assumed even for unlabeled images.
The method is designed for multi instance, single label datasets where an image is associated with multiple objects but only a single label. This work derives image labels through label propagation on the instance level.
In contrast, our method introduces label propagation for object detection datasets and leverages open-vocabulary models for region proposals. 

Tutevych et al.~\cite{tutevych2026} propose a framework for the semi-automated annotation of instance segmentation datasets.
A pretrained model is used for region proposals and proposed segments are embedded with multiple foundation models. The separation of classes in the embedding space is then learned using a contrastive loss. 
While this framework contains conceptually related ideas, it relies on a segmentation model pretrained on similar data and requires training for accurate label propagation. Instead, we focus on object detection datasets and do not require any training data besides the few labeled examples provided for label propagation.

\paragraph{Semi-Supervised Object Detection.}
While there is no prior work that applies graph-based SSL techniques in the domain of object detection, the research field of semi-supervised object detection is related to our approach. 
Methods in this field seek to improve detection performance by leveraging additional unlabeled data during training~\cite{papandreou2015,Jeong2019,sohn2020,liu2021,tang2021,chen2022denseSSL,Li2022,wang2023,soft_teacher}.
Typically, learning-based approaches are proposed that generate pseudo-labels based on few initial labels. 
These pseudo-labels can then be used as additional supervision during detector training which was shown to improve performance, compared to training with the limited labeled data available. 
Most existing approaches formulate semi-supervision at the image level and use detector-generated pseudo-labels rather than modeling similarities between object instances across the dataset.
In addition, approaches like CSD~\cite{Jeong2019}, STAC~\cite{sohn2020}, Humble Teacher~\cite{tang2021}, Consistent Teacher~\cite{wang2023} are learning-based, typically involving a student-teacher structure and data augmentations, which implies training overhead and high computational costs.
% Separating Paragraph
We develop graph-based SSL in the domain of object detection by leveraging open-vocabulary models which does not require training and offers explainability through appropriate quantification of both aleatoric and epistemic uncertainty of pseudo-labels. 

\paragraph{Open-Vocabulary Object Detection.}
Recent advances in open-vocabulary object detection enable localization of arbitrary object categories  through text prompts~\cite{minderer2022owlvit,liu2023grounding,cheng2024yoloworld}. 
Models such as Grounding DINO~\cite{liu2023grounding} have demonstrated strong zero shot performance and are increasingly used as building blocks for data annotation~\cite{zhao2024,DART}. 
In this work, we leverage open-vocabulary detectors to generate region proposals and object representations based on which we conduct graph-based label propagation. We compare our framework to the performance of these models alone.

\section{Method}
\label{sec: Method}

We introduce a graph-based semi-supervised framework for the uncertainty-aware annotation of object detection datasets. 
Given a small set of annotated objects, label information propagates to unlabeled object instances while both aleatoric and epistemic uncertainty are captured.
To this end, we leverage open-vocabulary models to generate object proposals and semantically meaningful object representations.
A similarity graph is constructed over these representations and probabilistic label spreading (PLS)~\cite{klees2026} is applied to diffuse information throughout the object-level graph. 
An overview of our method is illustrated in~\autoref{fig:method} where individual objects are iteratively queried for evaluation and received label information propagates. Importantly, our method also models background probabilities by including a distinct background class, which enables the separation of foreground objects.
The complete annotation framework consists of open-vocabulary model inference, post-hoc calibration of raw confidence and PLS to propagate available label information and quantify aleatoric and epistemic uncertainty of soft label estimates.

\paragraph{Task Description.}
The task at hand is the semi-automated annotation of object detection datasets with soft labels. 
We assume that a set of given images $I_1, \ldots, I_n$ contains a collection of objects $\mathcal{B}_1, \ldots, \mathcal{B}_n$ and each individual object $b_j\in \mathcal{B}_i$ is associated with a soft label \mbox{$p_j\in\Delta^{C+1}$}, where $C$ is the number of classes in the dataset and $\Delta^{C+1}$ denotes the $C+1$-dimensional probability simplex. 
The soft label is a categorical probability distribution over all classes including background, i.e., \mbox{$p_j = (p_j^1,\ldots,p_j^C, p_j^{bg})^\top$} such that $\sum_i p_j^i = 1$. 
It captures the aleatoric uncertainty associated with label assignment for this particular object, which is inherent to object detection datasets due to occlusions, blurring, small object size and ambiguous class affiliations~\cite{benchmark}. 
In this setting, the GT comprises all objects prevalent in the images and the associated soft labels are the underlying semantic distributions, i.e., we assume that manual class assignments for an object follow this distribution. In practice, annotated datasets do not cover all objects and, soft labels, if existing at all, are merely approximations to the underlying distributions based on finitely many annotations collected.

\paragraph{Object Proposals.}
We first apply an open-vocabulary object detector such as Grounding DINO~\cite{liu2023grounding} to all images and obtain a set of object proposals ${\hat{d}_1,\ldots,\hat{d}_M}$.
Each proposal $\hat{d}_j$ consists of a bounding box $\hat{b}_j$, an objectness score $\hat{s}_j\in[0,1]$, and an embedding vector $\hat{v}_j \in \mathbb{R}^{d}$ which encodes a representation of the detected object. 
Open-vocabulary models provide semantically rich representations that generalize across object categories and enable meaningful comparisons between object instances. We capture this information with the hidden representations of detected objects. Based on these embeddings, we construct a similarity graph for label propagation.

\paragraph{Post-hoc Calibration.}
Obtaining reliable pseudo-labels requires the accurate assessment of proposed objects. One signal we leverage is the detector confidence itself.
Since model confidences are often poorly calibrated~\cite{Chuan2017}, we conduct post-hoc calibration.
Based on the provided annotations, we  evaluate a subset of detections and calibrate model confidence using Isotonic Regression~\cite{isotonic_regression,isotonic_regression_2}.
For each evaluated detection, the target for model confidence is 1 if one of the object classes was correctly localized  or zero otherwise. 
We then apply the regression model to all predictions and obtain post-hoc calibrated confidences.
Ultimately, our framework blends the calibrated detector confidence in a Bayesian setting with independent score estimates obtained through label propagation. The set of pseudo-labels then results from choosing a suitable confidence threshold so that sufficiently many objects are annotated but the number of false positives is low.

\paragraph{Graph Construction.}
Based on the embeddings of proposed objects for the entire dataset, $\hat{v}_1, \ldots, \hat{v}_M$, we construct a similarity graph $G=(V,E,W)$, where each node corresponds to one object representation.
Edges are established between neighboring objects in the embedding space using a $k$-nearest-neighbor graph to ensure computational efficiency. The affinity matrix $W\in\RR^{M\times M}$ is built with a Gaussian kernel as in~\cite{Zhou_2003}, i.e.,
\begin{equation}
    W_{ij} = \exp\left(-\frac{1}{2\sigma^2} \| \hat{v}_i - \hat{v}_j\|^2\right),
\end{equation}
where $\sigma$ denotes the kernel's bandwidth parameter and a non-zero weight is only assigned to edges where at least one node is among the $k$-NN of the other. Diagonal entries of $W$ are set to zero to avoid self-reinforcement.
The affinity matrix captures semantic similarity between object instances across the entire dataset rather than within individual images and enables effective label propagation.

\paragraph{Probabilistic Label Spreading.}
From a semi-supervised perspective, out of the pool of region proposals $\hat{b}_1,\ldots,\hat{b}_M$ only a subset of objects is manually evaluated. We assume that each object $\hat{b}_j$ is associated with a probability distribution $\hat{p}_j \in \Delta^{C+1}$ so that querying an annotation for an object $\hat{b}_j$ can be interpreted as sampling a class $i_j\in\{1,\ldots,C,C+1\}$ from the underlying $\hat{p}_j$.
To propagate this label information, we employ Probabilistic Label Spreading (PLS)~\cite{klees2026}, a generalization of classical label spreading~\cite{Zhou_2003}. In PLS, label information is iteratively diffused through the graph according to the affinity matrix $W$ and Laplacian regularization. 
As proposed by~\cite{Zhou_2003}, we consider the normalized graph Laplacian 
\begin{equation}
    S = D^{-\frac{1}{2}} W D^{-\frac{1}{2}},
\end{equation}
based on which label information is distributed over the graph using the diffusion kernel
$(I-\alpha S)^{-1}$,
where $\alpha$ controls the strength of propagation.
More precisely, we encode the obtained annotation $i_j$ for object $\hat{p}_j$ in the initial label matrix $Y\in\RR^{M \times C+1}$ by setting $Y_{j,i_j} = 1$ and keeping all other entries zero.
This information then spreads as
\begin{equation}
    \label{eq:linear_system}
    F = (I-\alpha S)^{-1} Y .
\end{equation}
To approximate \eqref{eq:linear_system}, we use an efficient implementation of label spreading~\cite{AMELS} that constructs a sparse nearest neighbor graph and leverages algebraic multigrid solvers that often achieve near-linear complexity for such large sparse systems due to their multilevel structure~\cite{trottenberg2000multigrid}.
We normalize the propagation scores $F$ by dividing with the score of the labeled node, i.e., $\Tilde{F} = (F_{j,j})^{-1} F$ so that the labeled node receives a score of 1 and neighboring nodes receive partial label information. Hence, we can interpret propagation scores as the number of virtual labels accumulated in each node.
The propagation is repeated with each labeled object. Results are summed up and normalization yields estimated soft labels for every object proposal. Denoting $e_{i_j}\in\RR^{C+1}$ as the  unit vector resulting from the $j$-th query and having evaluated $m$ object proposals $\hat{b}_{1},\ldots, \hat{b}_{m}$, the soft label estimate for an object $\hat{b}_q$ is the following weighted average:
\begin{equation}
    \hat{p}_q = \frac{\sum_{j=1}^{m}   (I-\alpha S)^{-1}_{q,j}\, e_{i_j}}{\sum_{j=1}^{m}  (I-\alpha S)^{-1}_{q,j}} \in[0,1]^{C+1}
    %,\; q=1,\ldots,M\,
\end{equation}
PLS propagates probabilistic label information for individual objects and explicitly models uncertainty in the annotation process. As a result, based on few queried object labels, our method equips all proposed objects with estimated soft labels and captures epistemic uncertainty in terms of the absolute number of virtual labels received.

\paragraph{Background Modeling.}
Open-vocabulary detectors may generate false positive proposals. To account for this, we explicitly model the background class. During propagation, background information is treated analogously to semantic class information and contributes to the final probability estimates. This way, evaluated false positives enable the separation of foreground and background proposals.
Ultimately, each object proposal is assigned a soft label estimate
$\hat{p}_j = (\hat{p}_{j,1},\ldots,\hat{p}_{j,C},\hat{p}_{j,\mathrm{bg}})$
that separates foreground and background objects and captures class ambiguity while epistemic uncertainty can be derived from the number of virtual labels received.

Overall, our framework requires only a small number of manually annotated objects and does not involve any additional model training. Instead, it exploits the semantic structure induced by open-vocabulary object detectors to propagate single annotations across object detection datasets.

\newpage
%%%%%%%%%%%%%%%%%%%%%%%%%
% RESULTS
%%%%%%%%%%%%%%%%%%%%%%%%%

\newpage

\section{Numerical Results}
\label{sec: Results}

Our experiments demonstrate the effectiveness of our framework for the efficient and uncertainty-aware annotation of object detection datasets. We compare our method to baselines and similar algorithms and evaluate it on an established benchmark for semi-supervised object detection~\cite{sohn2020,soft_teacher}. We explore the effect of core components through ablation studies. We further reduce annotation costs through active label querying and demonstrate that our framework also uncovers label errors in existing datasets effectively.

\begin{itemize}[left=0pt]
    \item In \ref{subsec: Semi-automated Annotation}, we compare our method to baselines in terms of achieved data quality under limited annotation budget.
    \item To assess the effectiveness of our method, we compare it with state-of-the-art semi-supervised object detection methods on an established benchmark in~\ref{subsec: Semi-Supervised Object Detection}.
    \item Ablation studies in~\ref{subsec: Ablation Studies} provide an assessment of core components of our framework, in particular regarding the OVOD model employed and the feature space considered.
    \item We explore further reduction of annotation costs by leveraging active querying in~\ref{subsec: Active Querying}.
    \item As another application of our framework, we demonstrate its effectiveness regarding label error detection in~\ref{subsec: Label Error Detection}.
\end{itemize}

\subsection{Experimental Setup}
\label{subsec: Experimental Setup}

\paragraph{Datasets.} 
We evaluate our method
for uncertainty-aware annotation of object detection datasets 
on four datasets for which soft labels are available. In~\cite{benchmark}, common datasets were re-annotated from scratch with a focus on high coverage to also capture ambiguous cases. Identified objects were verified by at least 11 annotators, resulting in a soft label that quantifies human uncertainty.
Annotations are available for COCO~\cite{COCO} (validation split), Pascal VOC 2012~\cite{PascalVOC}, KITTI~\cite{kitti} (training split), and Cityscapes~\cite{Cityscapes}. If multiple splits are available, we consider them as a single dataset to be annotated at once.

\paragraph{Inference.}
We generate initial detections using Grounding DINO~\cite{liu2023grounding} in a zero-shot setting.
Following the official inference procedure, all categories are concatenated into a single text prompt, enabling simultaneous detection of all classes.
Inference results in 300 detections per image where we extract prediction confidence for each class. For YOLO-World~\cite{cheng2024yoloworld} and OWL-ViT~\cite{minderer2022owlvit} (considered in an ablation study), we follow the respective inference guidelines.

\paragraph{Label Spreading.}
To apply Probabilistic Label Spreading (PLS)~\cite{klees2026}, we set up a $k$-NN graph in the feature space of the ultimate hidden layer of the model's decoder. 
For Grounding DINO, these features have a dimension of 256. 
We include all predictions with a score of \gdinoscore{} or higher in the graph.
The bandwidth $\sigma$ of the Gaussian kernel is set to the average Euclidean $k$-NN distance of object features. To ensure efficient computation, as proposed by~\cite{AMELS}, we rely on Faiss~\cite{faisslibrary} for nearest neighbor search and use Algebraic Multigrid solvers with GPU acceleration~\cite{AMGX} to solve the linear system involved.

\paragraph{Annotation Budget.}
Our framework considers initial region proposals as the pool of unlabeled objects. 
We evaluate a fraction of these predictions regarding their class affiliation and propagate the resulting label information.
An annotation budget of 1\% signifies that accordingly many predictions are evaluated where sampling is conducted with replacement as we expect the label assignment to be probabilistic.
We simulate manual evaluation through IoU matching with ground truth (GT) using a threshold of $0.5$.
If a prediction does not match with any GT object, the background class is assigned, else, we sample a class from the corresponding GT soft label.

\paragraph{Baseline: Grounding DINO Pseudo-Labels.}
We compare our method to the baseline that uses the prediction confidence directly to include detections as pseudo-labels. Here, we keep all detections with a confidence of 0.5 or higher after post-hoc calibration.
Usually, confidence scores are only returned for the predicted class. To obtain a full soft label prediction, we extract confidence scores for the remaining classes as well and model the background probability as the product of counter-probabilities of all classes, i.e., $\hat{p}_{bg} = \prod_{\kappa=1}^C (1-\hat{p}_\kappa)$ where $\hat{p}_\kappa$ is the calibrated predicted probability for the $\kappa$-th class.

% RESULTS FIGURE

\begin{figure}[tb]
    \centering
    \includegraphics[width=\linewidth]{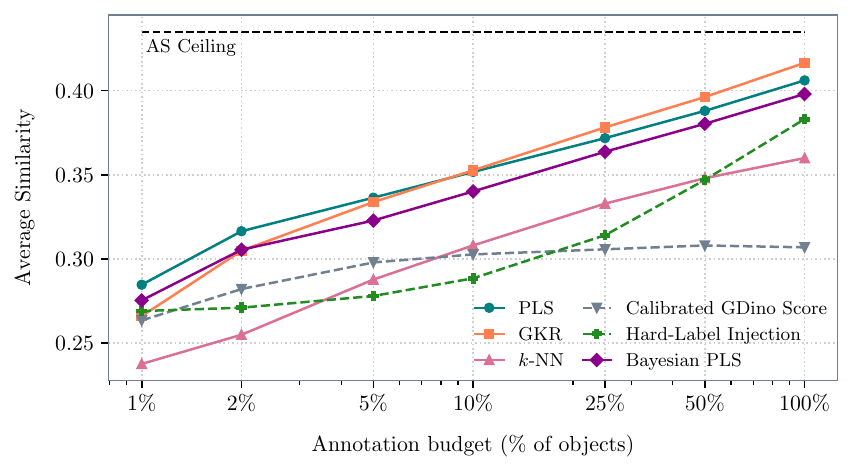}
    \caption{Average Similarity (AS\textsubscript{50:95})~\cite{benchmark} on the COCO dataset as a function of annotation budget for our propagation method (PLS)  against Gaussian Kernel Regression, $k$-NN regression, calibrated Grounding DINO confidence, and Hard-Label Injection. The dotted line marks the AS ceiling where class probabilities for all correctly localized detections perfectly align with the corresponding soft labels.}
    \label{fig:AS_comparison}
\end{figure}

\subsection{Semi-automated, Uncertainty-aware Annotation of Object Detection Datasets}
\label{subsec: Semi-automated Annotation}

% Experiment Scope
We compare our method with similar techniques and baselines. 
All methods assign soft labels to the initial region proposals based on the evaluated objects.
We assess the quality of the semi-automatically annotated dataset in terms of Average Similarity (AS)~\cite{benchmark}, a metric that generalizes Average Precision (AP) to the domain of soft labels by replacing the binary notion of precision with a normalized similarity measure to compare predicted probability distributions with soft labels. Following~\cite{benchmark}, we use 1~$-$~total variation distance as similarity measure, while similarity is zero for false positive predictions. 

Comparable methods are Gaussian Kernel Regression (GKR) \cite{Nadaraya1964,Watson1964} which propagates label information using a Gaussian Kernel, and $k$-NN regression where label information is averaged over the $k$-NN out of the labeled instances. 
For these parametrized methods, we conduct a grid search for hyperparameters and determine the best AS / AP value for each annotation budget for comparison. See appendix for more details and full parameter sweep results.
In addition, we consider two baselines. 
First, we calibrate model confidence using isotonic regression and infer soft label estimates from calibrated probabilities as described above. 
Second, we simulate fully manual annotation through hard-label injection. Here, we assign evaluated objects their labels but do not propagate information.
This yields a single annotation for evaluated objects and leaves non-evaluated objects at the soft label inferred from Grounding DINO scores.
 
As displayed in~\autoref{fig:AS_comparison}, on COCO validation data, our method achieves the highest AS at small annotation budgets (the regime of practical interest) and remains comparable to GKR as the budget grows further. Both clearly outperform calibrated Grounding DINO scores alone and hard-label injection across the full budget range, indicating that post-hoc calibration alone is insufficient to approximate soft label probabilities, and that manual hard-label annotation requires substantially more effort to derive high-quality soft labels. 
At high budgets, semi-supervised methods approach the AS ceiling. This upper bound is mainly due to the detection capabilities of Grounding DINO which does not detect each GT object and produces false positives. To compute this ceiling, we assign matching predictions the GT soft labels where an IoU threshold of 0.5 is used to match objects.

We evaluate achieved data quality in terms of AP in the appendix and find that our method surpasses Grounding DINO baseline performance with a moderate annotation budget ($5$--$10\%$).
This amount of label information seems to be needed to obtain accurate score estimates through graph-based label spreading.
A Bayesian blending of initial Grounding DINO score and our methods estimate leads to immediate improvements and performs comparably with hard-label injection that resorts predictions according to GT. Absolute gains in AP range from 1 percentage point at $10\%$ to $4$ at $100\%$ budget. 
To further investigate annotation quality, we compute precision and recall of COCO pseudo-labels for Bayesian PLS at $1\%$ annotation budget. We threshold annotations at the value that maximizes the $F_1$-score on labeled images. For large objects, this yields a precision of $88\%$ at a recall of $81\%$ meaning that the majority of large objects are correctly annotated.

We obtain similar results on Cityscapes and KITTI where semi-supervised methods outperform calibrated Grounding DINO and hard-label injection regarding AS. In terms of AP, however, hard-label injection is best on both datasets despite its comparatively low AS. This indicates that propagation-based methods better approximate soft labels while relative ranking of predictions is best with injection of available GT. On Pascal VOC, calibrated Grounding DINO is a strong baseline such that our method requires a larger annotation budget of around $25\%$ to overtake other methods in terms of both AS and AP. Interestingly, at a budget of 100\%, we even achieve a higher AP than hard-label injection meaning that we improve over the optimal ranking at the fixed IoU threshold of $0.5$ by smoothing label information over the graph. We discuss these dataset-specific findings further in the appendix.

% SECOND RESULTS FIGURE

\begin{figure}[tb]
    \centering
    \includegraphics[width=\linewidth]{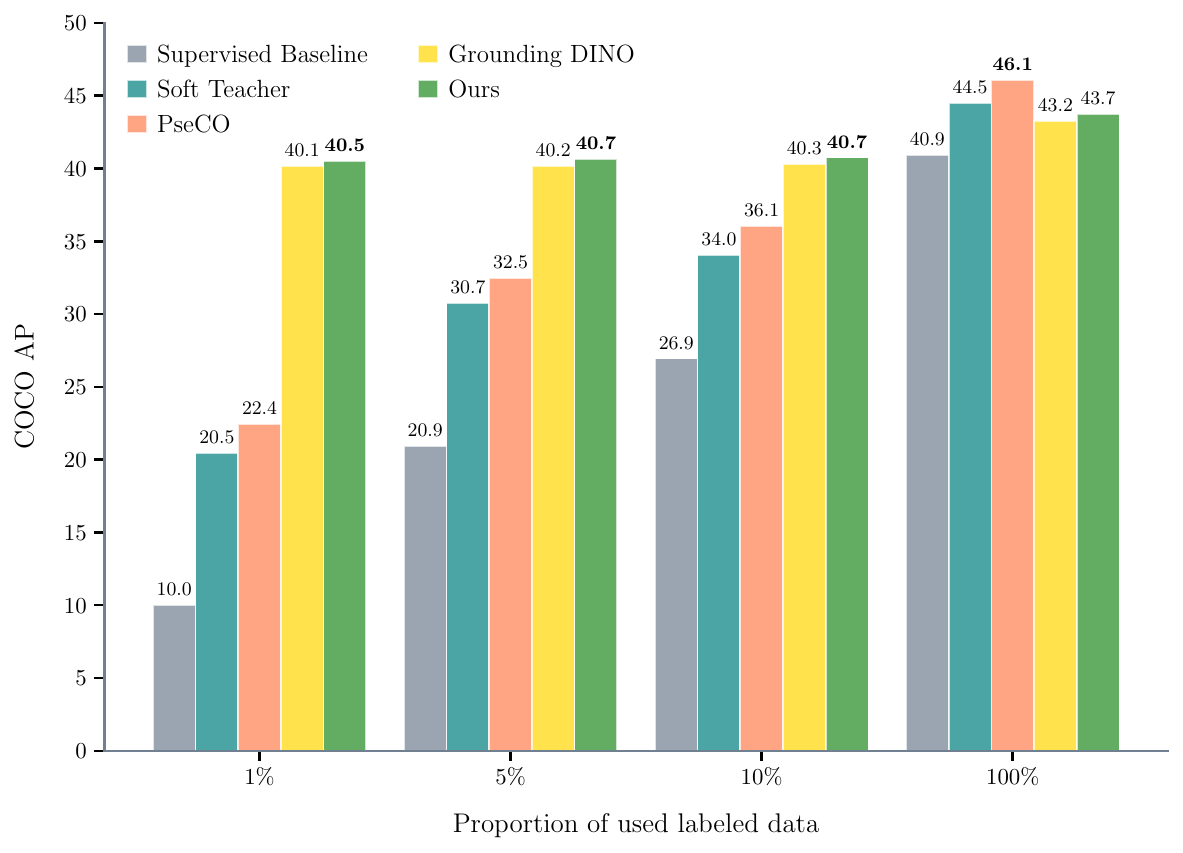}
    \caption{Benchmark results for semi-supervised object detection~\cite{sohn2020}. Average precision of a Faster R-CNN model with ResNet-50 backbone is reported on COCO validation data. The model is trained on pseudo-labels which are constructed using the label information available for a fraction of images. At 100\% budget, additional unlabeled COCO images are  used.}
    \label{fig:SSOD_benchmark_results}
\end{figure}

\subsection{Semi-Supervised Object Detection}
\label{subsec: Semi-Supervised Object Detection}

We compare our method to recent methods in the field of semi-supervised object detection (SSOD).
Here, label information is provided for a small fraction of images and the task is to generate pseudo-labels for the larger remainder of unlabeled images. Label quality is assessed by downstream detection performance of a Faster R-CNN model~\cite{fasterRCNN} with ResNet-50~\cite{ResNet} backbone. 
We rely on the implementation of Soft Teacher~\cite{soft_teacher} and follow the same training and evaluation protocol. Results on COCO validation data, averaged over the 5 folds of annotated training data is displayed in~\autoref{fig:SSOD_benchmark_results}, the corresponding standard deviations are listed in the appendix.

We significantly outperform existing methods already with minimal training data, achieving 40.5\% AP with 1\% of labeled images which almost matches the performance of the model trained on the entire training data (40.9\%).
This strong performance is due to the fact that, in contrast to existing learning-based methods, OVOD-based pseudo-labeling does not depend on training data since the model operates in a zero-shot setting.
The Grounding DINO baseline where annotations are used to calibrate model confidences in order to have a default choice of $0.5$ for the confidence threshold of pseudo-labels performs very well already (e.g., 40.1\% AP with 1\% of labeled data). It is further edged out by our Bayesian PLS approach that propagates received label information over the similarity graph to estimate independent scores and blends them with Grounding DINO scores. 
We optimize hyperparameter on the labeled data available as discussed in the appendix.

\begin{table}[t]
\centering
\caption{Downstream detection performance (AP in \%) for Faster R-CNN, YOLOv8-L, and RT-DETR trained with different annotation sets. We compare fully supervised training with Grounding DINO pseudo-labels and our pseudo-labeling approach, distinguishing two annotation budgets (1\%\ /\ 10\% of labeled images provided). Best (bold) and second best (underlined) values per row are highlighted.}
\label{tab:ssod_results_all_datasets}
\resizebox{\linewidth}{!}{%
\begin{tabular}{ll c cc cc}
\toprule
& & Full & \multicolumn{2}{c}{Grounding DINO} & \multicolumn{2}{c}{Ours} \\
\cmidrule(lr){4-5} \cmidrule(lr){6-7}
Dataset & Model & 100\% & 1\% & 10\% & 1\% & 10\% \\
\midrule
\multirow{3}{*}{COCO}
  & Faster R-CNN & 38.6 & 38.6 & 38.6 & \underline{38.7}   & \textbf{38.9}   \\
  & YOLOv8-L      & \underline{47.8} & 46.8 & 46.9 & 46.9 & \textbf{47.9}   \\
  & RT-DETR       & \textbf{47.7} & 46.7 & 46.4 & 46.7 & \underline{47.0}   \\
\addlinespace[2pt]
\multirow{3}{*}{VOC}
  & Faster R-CNN & 44.0 & 44.3 & \textbf{45.0} & \underline{44.8} & \underline{44.8} \\
  & YOLOv8-L      & 50.5 & 51.5 & 52.0 & \underline{52.8} & \textbf{53.1} \\
  & RT-DETR       & \textbf{66.4} & 65.3 & 65.4 & \underline{65.9} & \underline{65.9} \\
\addlinespace[2pt]
\multirow{3}{*}{Cityscapes}
  & Faster R-CNN & \textbf{33.7} & 25.6 & 27.1 & 19.6 & \underline{27.9} \\
  & YOLOv8-L      & \textbf{30.7} & 22.4 & 23.0 & 20.0 & \underline{23.7} \\
  & RT-DETR       & \textbf{38.7} & 27.7 & \underline{30.1} & 24.5 & 29.7 \\
\addlinespace[2pt]
\multirow{3}{*}{KITTI}
  & Faster R-CNN & \textbf{65.7} & 21.7 & 31.4 &  10.9 & \underline{32.6} \\
  & YOLOv8-L      & \textbf{77.7} & 20.9 & 29.4 & 12.7 & \underline{31.1} \\
  & RT-DETR       & \textbf{73.5} & 26.1 & \underline{34.6} & 15.5 & 34.5 \\
  \bottomrule
\end{tabular}
}
\end{table}

% \begin{table}[t]
% \centering
% \caption{Downstream detection performance (AP in \%) for Faster R-CNN and YOLOv8-L trained with different  annotation sets. We compare fully supervised training with Grounding DINO pseudo-labels and our pseudo-labeling approach, distinguishing two annotation budgets (1\%\ /\ 10\% of labeled images provided).}
% \label{tab:ssod_results_all_datasets}
% \resizebox{0.8\linewidth}{!}{%
% \begin{tabular}{ll c cc cc}
% \toprule
% & & Full & \multicolumn{2}{c}{Grounding DINO} & \multicolumn{2}{c}{Ours} \\
% \cmidrule(lr){4-5} \cmidrule(lr){6-7}
% Dataset & Model & 100\% & 1\% & 10\% & 1\% & 10\% \\
% \midrule
% \multirow{2}{*}{COCO}
%   & Faster R-CNN & 30.6 & \textbf{32.0} & 31.3 & 31.3   & 31.9   \\
%   & YOLOv8-L      & 35.6 & 35.6 & 35.9 & 35.7 & \textbf{36.0}   \\
% \addlinespace[2pt]
% \multirow{2}{*}{VOC}
%   & Faster R-CNN & 39.1 & 37.7 & 37.8 & 38.7 & \textbf{40.5} \\
%   & YOLOv8-L      & 49.2 & 48.7 & 49.4 & \textbf{50.3} & 49.8 \\
% \addlinespace[2pt]
% \multirow{2}{*}{Cityscapes}
%   & Faster R-CNN & \textbf{31.4} & 21.5 & 22.3 & 16.0 & 21.8 \\
%   & YOLOv8-L      & \textbf{28.6} & 22.1 & 22.4 & 20.4 & 22.4 \\
% \addlinespace[2pt]
% \multirow{2}{*}{KITTI}
%   & Faster R-CNN & \textbf{67.2} & 18.3 & 24.1 &  9.8 & 28.2 \\
%   & YOLOv8-L      & \textbf{73.1} & 24.4 & 31.4 & 14.2 & 33.3 \\
%   \bottomrule
% \end{tabular}
% }
% \end{table}

We assess downstream detection performance of Faster R-CNN~\cite{fasterRCNN}, YOLOv8-L~\cite{yolov8_ultralytics}, and RT-DETR~\cite{lv2023detrs} on four datasets in \autoref{tab:ssod_results_all_datasets}.
We follow the official training recipes and compare AP on validation data when training on the full set of annotations vs.\ constructed pseudo-labels.
Similar to COCO, we achieve comparable detection performance on Pascal VOC when training on pseudo-labels. In contrast, manual labels are superior for the driving scene datasets Cityscapes and KITTI where the gap in performance is most severe on KITTI. Our approach consistently performs better than the Grounding DINO baseline and overall, leads to competitive performance across models and datasets with limitations on complex street scene datasets, in particular, KITTI.

Overall, we significantly push the state of the art in SSOD, virtually matching detection performance of the fully supervised model on COCO and Pascal VOC with very little labeled data. 
These results underline the effectiveness of current open-vocabulary models and their usefulness for semi-automated annotation. 
Given that our method closely approaches the 'ceiling' of the fully supervised model, it should be noted that the evaluation protocol is biased towards the original dataset. These annotations are treated as the gold standard even though manual annotations have limitations as well~\cite{benchmark}. While it is not possible to directly compare data quality between two sets of annotations (each of them being imperfect), accurate assessment of downstream detection performance requires a third, independent set of annotations.

With the refined annotations of~\cite{benchmark}, we can re-evaluate the performance of models on independent annotations (\autoref{tab:ssod_results_all_datasets_softgt}). 
The refined annotations are more rigorous and include more object annotations than the original ones. 
For COCO, Pascal VOC, and Cityscapes, a similar pattern as in \autoref{tab:ssod_results_all_datasets} can be observed. 
Pseudo-labeling leads to overall competitive performance, where manual annotation remains best on Cityscapes, while pseudo-labeling is best on Pascal VOC and results are mixed on COCO. Results on KITTI, however, change significantly. 
Models trained on pseudo-labels were evaluated as significantly worse on original annotations but actually score better on refined annotations, compared to models trained on the original data. This highlights the dependence on evaluation data, where YOLOv8-L, e.g., drops from 78\% AP on original labels to only 22\% when evaluating on the refined annotations. 

\begin{table}[tb]
\centering
\caption{Ablation Results: Adding additional models to Grounding DINO as an ensemble does not improve recall of COCO pseudo-labels but introduces false positives (left). Embedding detected objects with CLIP~\cite{CLIP} for graph construction leads to overall worse results (right).}
\label{tab:ablation_ovod_clip}
\begin{subtable}{0.48\linewidth}
\setlength{\tabcolsep}{3pt}
\centering
\caption{OVOD Ensemble where predictions are pooled using NMS.}
\label{tab:ovod_ensemble_ablation}
\resizebox{\linewidth}{!}{%
\begin{tabular}{lrrr}
\toprule
Method & Prec. & Recall & F1 \\
\midrule
Grounding DINO & 84.9 & 61.4 & \textbf{71.3} \\
+ YOLO-World & 84.6 & 61.5 & 71.2 \\
+ YOLO-World + OWL-ViT & 82.7 & 61.5 & 70.5 \\
\bottomrule
\end{tabular}
}
\end{subtable}
\hfill
\begin{subtable}{0.48\linewidth}
\setlength{\tabcolsep}{3pt}
\centering
\caption{Grounding DINO vs.\ CLIP features for graph construction.}
\label{tab:clip_graph_ablation}
\resizebox{\linewidth}{!}{%
\begin{tabular}{l|rrr|rrr}
\toprule
 & \multicolumn{3}{c}{AP} & \multicolumn{3}{c}{AS} \\
Variant & 1\% & 10\% & 100\% & 1\% & 10\% & 100\% \\
\midrule
GDINO & \textbf{44.4} & \textbf{48.6} & \textbf{52.7} & \textbf{27.8} & \textbf{31.8} & \textbf{34.9} \\
CLIP & 29.4 & 35.8 & 49.9 & 10.8 & 17.1 & 21.5 \\
% CLIP + prior & 49.3 & 49.9 & 52.5 & 24.8 & 29.9 & 32.9 \\
% GDINO + prior & 49.3 & 50.3 & 52.7 & 27.2 & 32.0 & 34.9 \\
% GDINO + CLIP + prior & 49.4 & 50.2 & 53.5 & 25.6 & 31.3 & 34.7 \\
\bottomrule
\end{tabular}
}
\end{subtable}
\end{table}

\subsection{Ablation Studies}
\label{subsec: Ablation Studies}

Our framework has two main design choices. For one, we chose Grounding DINO rather than a different or an ensemble of open-vocabulary models. Second, we build the similarity
graph on Grounding DINO embeddings instead of using a separate model such as CLIP~\cite{CLIP}. These choices are supported by two ablation studies, summarized in \autoref{tab:ablation_ovod_clip}.

First, we compare Grounding DINO alone against two OVOD ensembles (adding
YOLO-World~\cite{cheng2024yoloworld}, and additionally OWL-ViT~\cite{minderer2022owlvit}). We first calibrate models individually and retain predictions with calibrated confidence of at least $0.5$. To pool predictions, we apply NMS with an IoU threshold of $0.5$, where in case of overlap, Grounding DINO boxes are kept. 
We measure pseudo-label quality in terms of recall and precision on COCO training data at an IoU threshold of $0.5$.
Recall remains unchanged indicating that the other models do not detect any GT objects that Grounding DINO missed. However, they introduce false positives so that precision drops slightly and Grounding DINO alone achieves the highest $F_1$ score. 

Second, we compare performance for two similarity graphs. One from Grounding DINO's own
decoder embeddings and one built from CLIP embeddings of the
same detections. 
The CLIP graph leads to poorer performance, particularly for small budgets. 
We therefore build the graph on Grounding DINO's embeddings directly. It is the simpler design and consistently performs best on both metrics.

\subsection{Active Querying of Objects to Annotate}
\label{subsec: Active Querying}

\begin{figure}[t]
    \centering
    \includegraphics[width=\linewidth]{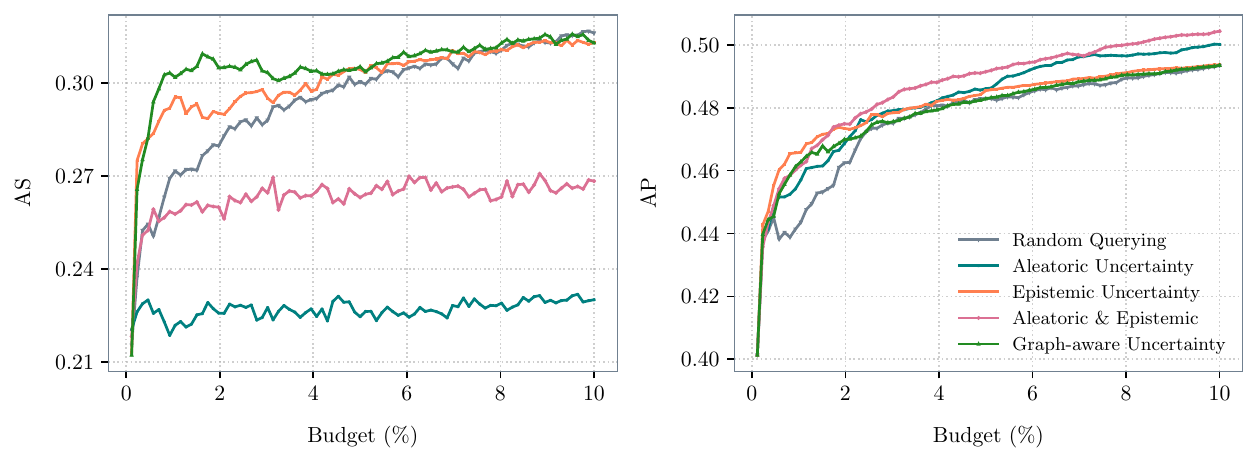}
    \caption{Active querying results: We compare performance in terms of AS (left) and AP (right) on COCO for five querying strategies that choose different nodes to annotate next.}
    \label{fig:AL_results}
\end{figure}

From a practical perspective, we aim to further reduce annotation effort by actively querying labels.
On COCO validation data, we query one label at a time and evaluate achieved AS and AP every 100 queries. With each queried label, the soft label predictions are updated and we can assess aleatoric and epistemic uncertainty in every graph node.
We compare four strategies to the baseline of random querying. 
Querying by aleatoric uncertainty which is quantified as the Shannon entropy of current soft label estimates to focus on ambiguous instances. 
Querying by epistemic uncertainty, quantified by the number of virtual labels received in each node to annotate the node that received the least label information so far. 
An equally weighted mixture of the two sources of uncertainty, where we normalize quantities using the maximal and minimal value occurring.
And lastly, we consider a strategy that combines aleatoric and epistemic uncertainty with the structure of the graph. It considers the width of the 95\% Wilson score interval ~\cite{Wilson1927} (summed over all classes) based on the virtual labels received in each node and smoothens this quantity using the heat kernel $(I-\alpha S)^{-1}$. More precisely, if $w_i$ denotes the interval width for node $i$, summed over all classes, the node to be queried next is the one that maximizes
\begin{equation}
    s_j = \sum_{i=1}^{n}(I-\alpha S)^{-1}_{ij}\, w_i \Bigm/ \sum_{i=1}^{n} (I-\alpha S)^{-1}_{ij}.
\end{equation}
Overall, the results in \autoref{fig:AL_results} show that active querying can further reduce the annotation effort required to achieve a certain degree of data quality.
Querying by epistemic uncertainty ensures broad coverage of the feature space, improving both metrics over the random baseline. Meanwhile, querying by aleatoric uncertainty may overly focus on boundary regions where ambiguous instances reside, reducing the quality of soft labels through noisy labels. However, the mixture of both sources of uncertainty shows best AP results while querying based on Wilson intervals and the graph structure leads to best AS results. 
While AS and AP behave quite differently, it can be concluded that both aleatoric and epistemic uncertainty are relevant for active querying.

\subsection{Label Error Detection via Label Smoothing}
\label{subsec: Label Error Detection}

Using the label error evaluation protocol of~\cite{benchmark}, we assess the effectiveness of our framework for detecting label errors such as missing or poorly localized annotations.
Our method smoothens all GT labels over the graph of detected objects such that false positive predictions that are semantically similar to true positives are considered as label error candidates. 
The pool of label error proposals are NMS-filtered Grounding DINO predictions (IoU$\geq 0.6$) with a minimal score of 0.01. We compare ranking quality of these proposals between a random sorting baseline, Grounding DINO and Bayesian PLS. Hyperparameters of Bayesian PLS are optimized towards AP on held-out data using only original annotations. 
Results are listed in \autoref{tab:label_errors}. Bayesian PLS is best on COCO and Cityscapes while Grounding DINO alone is best on KITTI and Pascal VOC. Notably, our approach is best on Cityscapes where few label errors exist and worst on KITTI, where original and refined annotations are quite different. Since label propagation increases bias towards GT, it can improve label error detection when annotations are mostly complete but inherits a bias when original and refined annotations diverge.

% \begin{table}[t]
% \centering
% \caption{Comparison of label error detection capabilities.
% % Non-matching predictions are considered as label error proposals and compared to a set of actual label errors for evaluation. A label error proposal is considered correct if it matches a real label error with an IoU $\geq 0.1$.
% We report AP and the best F1 score based on each method's own precision/recall curve.}
% \label{tab:label_errors}
% \begin{tabular}{llrr}
% \toprule
% Dataset & Method & AP & Best F1  \\
% \midrule
% COCO & GDino  & 27.32 & 39.69 \\
%   & Bayesian PLS & 27.60 & 40.31 \\
% Pascal VOC & GDino & 34.00 & 41.90  \\
%   & Bayesian PLS  & 33.01 & 40.09  \\
% \bottomrule
% \end{tabular}
% \end{table}

\begin{table}[t]
\centering
\caption{Comparison of label error detection capabilities.
We consider originally false positive predictions as label error proposals and evaluate them on actual label errors using the evaluation protocol of~\cite{benchmark} that reports AP and maximal $F_1$. 
%'s precision-recall curve (bold indicates best).
}
\label{tab:label_errors}
\resizebox{\linewidth}{!}{%
\begin{tabular}{lrrrrrrrr}
\toprule
 & \multicolumn{2}{c}{COCO} & \multicolumn{2}{c}{Cityscapes} & \multicolumn{2}{c}{KITTI} & \multicolumn{2}{c}{Pascal VOC} \\
\cmidrule(lr){2-3} \cmidrule(lr){4-5} \cmidrule(lr){6-7} \cmidrule(lr){8-9}
Method & AP & F1 & AP & F1 & AP & F1 & AP & F1 \\
\midrule
Random & 3.50 & 7.30 & 1.69 & 3.36 & 3.50 & 6.76 & 0.65 & 1.30 \\
GDino & \underline{34.51} & \underline{40.89} & \underline{19.36} & \textbf{26.99} & \textbf{58.80} & \textbf{57.05} & \textbf{35.91} & \textbf{41.29} \\
Bayesian PLS & \textbf{34.56} & \textbf{41.16} & \textbf{21.03} & \underline{26.66} & \underline{37.62} & \underline{41.96} & \underline{33.98} & \underline{37.88} \\
\bottomrule
\end{tabular}%
}
% \begin{tabular}{lrrrrrrrr}
% \toprule
%  & \multicolumn{2}{c}{COCO} & \multicolumn{2}{c}{Cityscapes} & \multicolumn{2}{c}{KITTI} & \multicolumn{2}{c}{Pascal VOC} \\
% \cmidrule(lr){2-3} \cmidrule(lr){4-5} \cmidrule(lr){6-7} \cmidrule(lr){8-9}
% Method & AP & F1 & AP & F1 & AP & F1 & AP & F1 \\
% \midrule
% Random & 3.52 & 7.35 & 1.76 & 3.49 & 3.40 & 6.58 & 0.67 & 1.33 \\
% Grounding Dino & \underline{34.99} & \underline{41.25} & \underline{20.01} & \underline{27.57} & \textbf{58.13} & \textbf{56.28} & \textbf{37.05} & \textbf{42.07} \\
% % PLS & 10.59 & 19.83 & 4.17 & 8.17 & 11.49 & 21.49 & 2.83 & 5.46 \\
% Bayesian PLS & \textbf{35.03} & \textbf{41.51} & \textbf{21.75} & \textbf{27.58} & \underline{36.32} & \underline{41.12} & \underline{35.01} & \underline{38.58} \\
% \bottomrule
% \end{tabular}
% }
\end{table}

%%%%%%%%%%%%%%%%%%%%%%%%%
% CONCLUSION
%%%%%%%%%%%%%%%%%%%%%%%%%

\section{Conclusion}
We proposed a framework for graph-based semi-supervised learning in object detection. It enables the semi-automated, uncertainty-aware annotation of object detection datasets with soft labels.
We demonstrated its effectiveness by comparing it to baselines in terms of achieved soft label quality on multiple soft-labeled datasets.
We showed that semi-automatically constructed COCO annotations are of high quality as training on them leads to virtually matching downstream detection performance, compared to training on manually constructed data. This significantly improves the state of the art for semi-supervised object detection in the regime of partially labeled COCO training data.
We further demonstrated that data-centric evaluation heavily depends on the evaluation data and treating manual annotations as the gold standard is an implicit assumption that may limit the accurate assessment of pseudo-label quality.
We integrated active querying techniques and demonstrated their potential to further reduce annotation effort.
Finally, we showed that our framework effectively uncovers label errors in existing datasets by smoothing labels over the object-level similarity graph, achieving state-of-the-art performance on two out of four datasets.

To support reproducibility and further research, we make our source code publicly available under \url{https://github.com/JonathanKlees/Uncertainty-aware-Pseudo-Labeling}.

\section*{Acknowledgements}
J.K.\ and M.R.\ acknowledge support by the German Federal Ministry of Research, Technology and Space (BMFTR) within the project RELiABEL (grant no.\ 16IS24019B).

% References
\bibliographystyle{IEEEtran}
\bibliography{literature}

\newpage

\renewcommand{\thefigure}{A.\arabic{figure}} 
\renewcommand{\thetable}{A.\arabic{table}}   
\setcounter{figure}{0}                        
\setcounter{table}{0}                         

\appendix
\label{sec:Appendix}
% Appendix headings should be subsections
\newcommand{\appendixheading}{\subsection}
\appendixheading{Supplementary Material Overview}
This appendix complements the main manuscript with additional empirical studies, implementation details, and visual examples, which were beyond the scope of the main paper. Below, we give a brief overview to facilitate the navigation of the supplementary material.
\begin{itemize}[left=0pt]
\item Section~\ref{sec-appdx:additional_ssod_results} lists the full results and hyperparameters for the Semi-supervised Object Detection benchmark. These results are illustrated as a bar chart in the main paper and are now reported with two decimal places and standard deviations with a comprehensive comparison to other methods.
\item Section~\ref{sec:appendix-annotation_results} complements the implementation details and results for achieved annotation quality of semi-supervised methods and baselines with average precision as a second metric and performance on other datasets.
\item Section~\ref{sec: Appendix - Hyperparameter Study} studies the effect of hyperparameters such as the spreading intensity, the number of neighbors and the minimal confidence for initial predictions.
\item Section~\ref{sec:appendix-active_querying} provides more details regarding the active querying techniques applied and provides additional insights regarding uncertainty-based querying.
\item Section~\ref{sec: Appendix - Annotation Costs} compares annotation costs of our semi-automated approach with fully manual dataset construction of soft-labeled datasets.
\item Section~\ref{sec: Appendix - Visual Examples} provides visual examples of detected objects and compares soft labels with the estimates of Grounding DINO alone and our framework.
\end{itemize}

\appendixheading{Full Results for Semi-Supervised Object Detection}
\label{sec-appdx:additional_ssod_results}

\begin{table}[htbp]
\centering
\resizebox{\linewidth}{!}{
\begin{tabular}{llllc}
\toprule
Method & 1\% & 5\% & 10\% & 100\% \\
\midrule
Supervised Baseline & 10.00 $\pm$ 0.26 & 20.92 $\pm$ 0.15 & 26.94 $\pm$ 0.11 & 40.90 \\
CSD~\cite{Jeong2019} & 10.51 & 18.63 & 22.46 & 38.82 \\
STAC~\cite{sohn2020} & 13.97 $\pm$ 0.35 & 24.38 $\pm$ 0.12 & 28.64 $\pm$ 0.21 & 39.21 \\
Instant Teaching~\cite{zhou2021instant} & 18.05 & 26.75 & 30.40 & - \\
Humble Teacher~\cite{tang2021} & 16.96 & 27.70 & 31.61 & - \\
Unbiased Teacher~\cite{liu2021} & 20.75 & 28.27 & 31.50 & 41.30 \\
Soft Teacher~\cite{soft_teacher} & 20.46 $\pm$ 0.39 & 30.74 $\pm$ 0.08 & 34.04 $\pm$ 0.14 & 44.50 \\
ACRST~\cite{zhang2021semi} & 26.07 & 31.35 & 34.92 & 42.79 \\
PseCO~\cite{li2022pseco} & 22.43 & 32.50 & 36.06 & \textbf{46.10} \\
Grounding DINO~\cite{liu2023grounding} & 40.14 $\pm$ 0.11 & 40.19 $\pm$ 0.08 & 40.30 $\pm$ 0.10 & 43.24 \\
Ours & \textbf{40.50} $\pm$ 0.23 & \textbf{40.65} $\pm$ 0.11 & \textbf{40.74} $\pm$ 0.08 & 43.71 \\
\bottomrule
\end{tabular}
}
\caption{
    Comparison of different semi-supervised object detection methods on COCO.
    The results are reported for different fractions of annotated training data in terms of mean average precision (AP).
    For annotation budgets below 100\%, results are averaged over 5 folds of annotated data and standard deviation is reported, following the evaluation protocol of~\cite{sohn2020}.
    We list the numbers for existing methods as reported by~\cite{soft_teacher} and~\cite{wang2023} where standard deviations are not always given.
    }
\label{tab:ssod_results_coco}
\end{table}

In~\autoref{tab:ssod_results_coco}, the full results for the semi-supervised object detection benchmark are collected. We follow the evaluation protocol proposed by~\cite{sohn2020}, which is also implemented in the codebase of~\cite{soft_teacher}. We list the results of existing methods as reported by~\cite{soft_teacher} and~\cite{wang2023} and apply the same training schemes with the pseudo-labels we constructed to generate the results for our method and the Grounding DINO baseline.
The proposed training scheme for the Faster R-CNN~\cite{fasterRCNN} with ResNet-50~\cite{ResNet} backbone and Feature Pyramid Network~\cite{lin2017featurepyramidnetworks} uses an initial learning rate of $0.01$, which is divided by $10$ after $120,000$ and once again after $160,000$ iterations. The training runs for $180,000$ iterations in total on $8$ GPUs with a batch size of $5$ images per GPU. In contrast, for the budget of $100\%$, the training schedule is four times longer ($720,000$ iterations with the above learning rate decays at $480,000$ and $640,000$ iterations) and the batch size is increased to $8$. Further training parameters are based on the MMDetection framework~\cite{MMDetection} and we refer to the supplementary material of~\cite{soft_teacher} for the full description of hyperparameters and further training details.
We computed on 8 NVIDIA RTX6000 Pro Backwell GPUs with 96GB of memory each where one training of $180,000$ iterations takes about 10h and full benchmark evaluation takes about 9 days per method.

We generate pseudo-labels with our Bayesian PLS method that blends the initial Grounding DINO with the PLS derived propagation score. More precisely, if $\hat{p}_j^{GD}$ denotes the soft label estimate by Grounding DINO after post-hoc calibration and $F_j\in\mathbb{R}^{C+1}$ denotes the raw propagation scores derived from PLS in the $j$-th node, the Bayesian PLS estimate is
\begin{equation}
    \hat{p}_j = \kappa\ \hat{p}_j^{GD} + F_j,
\end{equation}
where $\kappa > 0$ is the prior weight and $F_j$ is a vector that corresponds to the virtual number of labels received for each class.
We use the following procedure for pseudo-label construction. For the budgets $1\%, 5\%, 10\%$, we first build the graph of Grounding DINO embeddings of all predictions with a minimal confidence of $0.2$ on the entire COCO training dataset. Based on the images that serve for supervision, we evaluate the corresponding predictions using an IoU threshold of $0.5$ and non-greedy matching. We then propagate the binary information (TP / FP) across the graph to obtain independent score estimates for all objects, which are then blended with the calibrated Grounding DINO score. We fit an isotonic regression model to calibrate Grounding DINO confidence using the TP / FP evaluations of the predictions on the labeled images as binary targets. We optimize the diffusion hyperparameters with a grid search of $k \in \{10, 20, 50\}$, $\alpha \in \{0.7, 0.85, 0.9, 0.95, 0.99\}$ and the Bayesian blend weight $\kappa \in \{0.01, 0.03, 0.05, 0.1, 0.2, 0.3, 0.5, 1.0, 2.0, 5.0, 10.0\}$ using an 80/20 image-level split of the provided labeled images. We select the parameter combination that maximizes the $F_1$-score on the 20\% of held-out images by propagating the label information from the remaining 80\% of labeled images. Hence, the best operating point in terms of $F_1$-score on the respective precision-recall curve is determined and this optimal threshold is then used to include pseudo-labels. 
To derive the final set of pseudo-labels, the model with the optimal parameter combination is then evaluated using the full set of provided labeled images as supervision for label propagation.

\begin{figure}[tb]
    \centering
    \includegraphics[width=\linewidth]{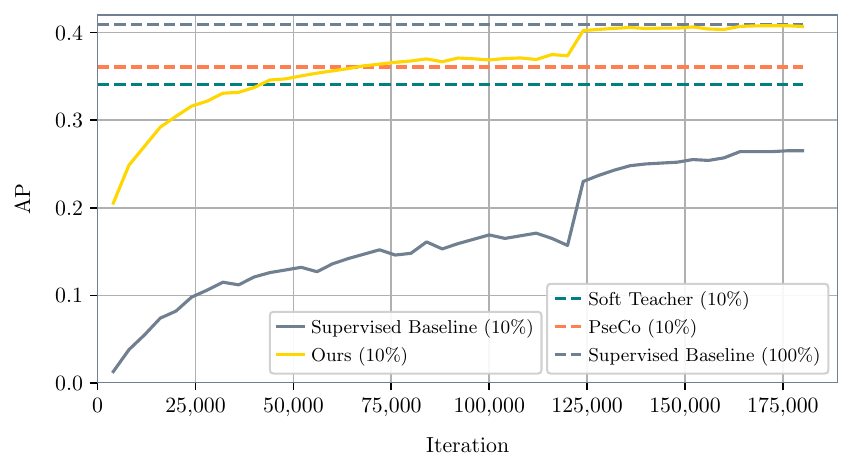}
    \caption{Training curves for Faster R-CNN with ResNet-50 backbone on COCO for the supervised baseline with $10\%$ of labeled images and our pseudo-labels. We observe a fast increase in validation AP that ultimately approaches the performance of the model trained on the entire set of training data and surpasses other pseudo-labeling approaches (dashed lines).}
    \label{fig:SSOD_training_curves}
\end{figure}

Comparing the training curves in \autoref{fig:SSOD_training_curves} that display validation AP depending on the number of iterations, we observe that the Faster R-CNN model learns significantly faster due to the additional pseudo-labeled images available, compared to the supervised baseline. Here, we consider a budget of $10\%$ and the first of five folds of labeled images provided for benchmark evaluation~\cite{soft_teacher}. For this budget and fold, we ultimately match the performance of the model that is trained on the full training data, which indicates the high quality of generated pseudo-labels. In particular, since the baseline model leverages a four times longer training schedule with a larger batch size as well, so images are processed more often during training and gradients may be more precise.

To assess downstream detection performance on all four datasets, we train Faster R-CNN~\cite{fasterRCNN}, YOLOv8-L~\cite{yolov8_ultralytics}, and RT-DETR~\cite{lv2023detrs} once on the full training data and in comparison, on pseudo-labeled data. Pseudo-labels are based on either 1\% or 10\% of annotated images and generated with Grounding DINO alone or Bayesian PLS as described above.
All three models are trained with standard hyperparameters regarding optimizer, learning-rate schedule and augmentations and we deviate only as described below.
Every model is initialized from ImageNet-pretrained backbone weights not from a COCO-pretrained checkpoint so that the comparison is not skewed by an existing COCO prior. We follow each architecture's official training recipe where YOLOv8-L is trained for 500 epochs and Faster R-CNN for 26 epochs (the latter distributed across 4 GPUs to match the reference batch size) on all four datasets, while RT-DETR is trained for 100 epochs on Pascal VOC, Cityscapes and KITTI and for 10 epochs on COCO. All models are scored in terms of AP on each dataset's validation split.

To reduce the dependence on original annotations, we also evaluate all models on independently constructed annotations for all datasets from~\cite{benchmark}. 
The motivation for this is that the assessment of downstream detection performance assumes that original annotations are the gold standard by which model capabilities are to be measured. That way, the model that is trained on the most similar distribution of annotations (i.e, the original annotations if we assume the distribution is the same across training and validation data) is scored the highest in evaluation. But it might be that these annotations are imperfect and a better set of labels may exist. Hence, we compare annotation quality in terms of downstream detection performance on independent data to evaluate how well a model generalizes to these independent annotations based on the training data it was provided. 
The results in terms of average precision on validation data are displayed in \autoref{tab:ssod_results_all_datasets_softgt}. 

\begin{table}[t]
\centering
\caption{Downstream detection performance (AP in \%) for Faster R-CNN, YOLOv8-L, and RT-DETR, on independently constructed annotations of each dataset's validation set, which were found to cover more object annotations~\cite{benchmark}.}
\label{tab:ssod_results_all_datasets_softgt}
\resizebox{\linewidth}{!}{%
\begin{tabular}{ll c cc cc}
\toprule
& & Full & \multicolumn{2}{c}{Grounding DINO} & \multicolumn{2}{c}{Ours} \\
\cmidrule(lr){4-5} \cmidrule(lr){6-7}
Dataset & Model & 100\% & 1\% & 10\% & 1\% & 10\% \\
\midrule
\multirow{3}{*}{COCO}
  & Faster R-CNN & 33.9 & \underline{34.1} & \underline{34.1} & \underline{34.1} & \textbf{34.4} \\
  & YOLOv8-L      & \textbf{44.5} & 42.9 & 43.1 & 42.9 & \underline{43.6} \\
  & RT-DETR       & \textbf{42.7} & 42.0 & 41.7 & 42.1 & \underline{42.2} \\
\addlinespace[2pt]
\multirow{3}{*}{VOC}
  & Faster R-CNN & 37.8 & 38.2 & 38.8 & \textbf{39.9} & \underline{39.0} \\
  & YOLOv8-L      & 45.4 & 46.0 & 46.5 & \textbf{48.2} & \underline{46.8} \\
  & RT-DETR       & 57.4 & 57.1 & 57.6 & \textbf{58.8} & \underline{57.9} \\
\addlinespace[2pt]
\multirow{3}{*}{Cityscapes}
  & Faster R-CNN & \textbf{35.5} & 28.4 & 30.0 & 21.8 & \underline{31.5} \\
  & YOLOv8-L      & \textbf{33.1} & 25.5 & 25.3 & 21.7 & \underline{27.4} \\
  & RT-DETR       & \textbf{40.7} & 31.6 & \underline{33.3} & 26.8 & 33.2 \\
\addlinespace[2pt]
\multirow{3}{*}{KITTI}
  & Faster R-CNN & 22.8 & 19.3 & \underline{23.0} & 10.3 & \textbf{25.8} \\
  & YOLOv8-L      & 21.7 & 23.3 & \underline{24.8} & 12.0 & \textbf{28.2} \\
  & RT-DETR       & 24.3 & 26.1 & \underline{29.6} & 14.3 & \textbf{30.1} \\
  \bottomrule
\end{tabular}
}
\end{table}

\appendixheading{Further Results on Semi-automated, Uncertainty-aware Dataset Annotation}
\label{sec:appendix-annotation_results}

\begin{figure*}[tb]
    \centering
    \begin{subfigure}[t]{0.32\linewidth}
        \includegraphics[width=\linewidth]{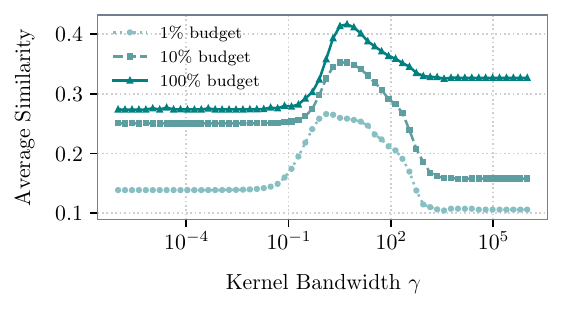}
        \caption{Gaussian Kernel Regression}
    \end{subfigure}
    \hfill    
    \begin{subfigure}[t]{0.32\linewidth}
        \includegraphics[width=\linewidth]{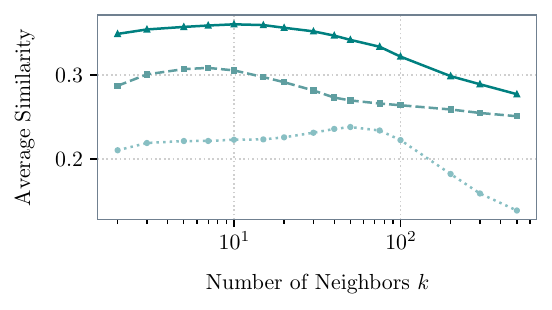}
        \caption{$k$-NN Regression}
    \end{subfigure}
    \hfill    
    \begin{subfigure}[t]{0.32\linewidth}
        \includegraphics[width=\linewidth]{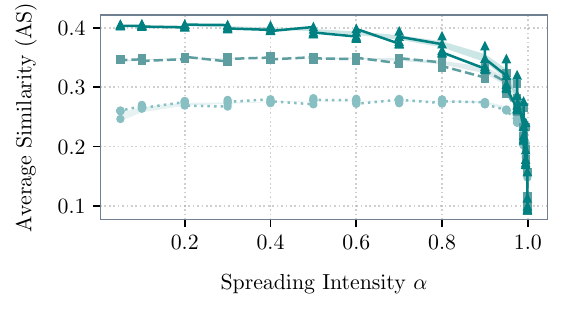}
        \caption{PLS}
    \end{subfigure}
    
    \caption{Performance in terms of Average Similarity for parametrized methods depending on hyperparameter values. Hyperparameters were evaluated over a grid for each annotation budget. For readability, we display the three annotation budgets $1\%$, $10\%$, and $100\%$. For PLS, since there are two hyperparameters, we indicate dependence on the parameter $k$ by shading the minimal and maximal obtained value for each spreading intensity $\alpha$.}
    \label{fig:full_sweep_results}
\end{figure*}

\paragraph{Implementation Details.}
In the main manuscript we report achieved Average Similarity with GT on the COCO dataset as a function of the annotation budget provided. This budget indicates the fraction of pre-filtered detections evaluated (we keep those with confidence of at least 0.2) and we list the optimal value for each budget after conducting a hyperparameter grid search for parametrized methods. 
PLS is grid-searched over $k \in \{10, 20\}$, a kernel-bandwidth scale $\gamma\in \{0.1, 0.2, 0.5, 1.0, 2.0\}$ which scales the  Gaussian kernel width, relative to the mean squared $k$-NN distance used by default and $\alpha$ at 15 values spanning $0.05$–$0.999$. The Bayesian PLS builds on the winning parameter combination and sweeps the Grounding DINO prior weight $\kappa$ (15 values, $0.1$–$10^5$). Gaussian kernel regression has a single parameter, the kernel bandwidth $\gamma$, which is swept across 60 log-spaced values, $10^{-6}$–$10^6$). $k$-NN regression is evaluated for 15 $k$ values ranging from 2–500), while the calibrated Grounding DINO baseline and hard label injection need no hyperparameter search. The sweep results for parametrized methods are displayed in~\autoref{fig:full_sweep_results}, where it can be seen that the parameter spaces have been swept properly for all methods.

\begin{figure}[tb]
    \centering
    \includegraphics[width=\linewidth]{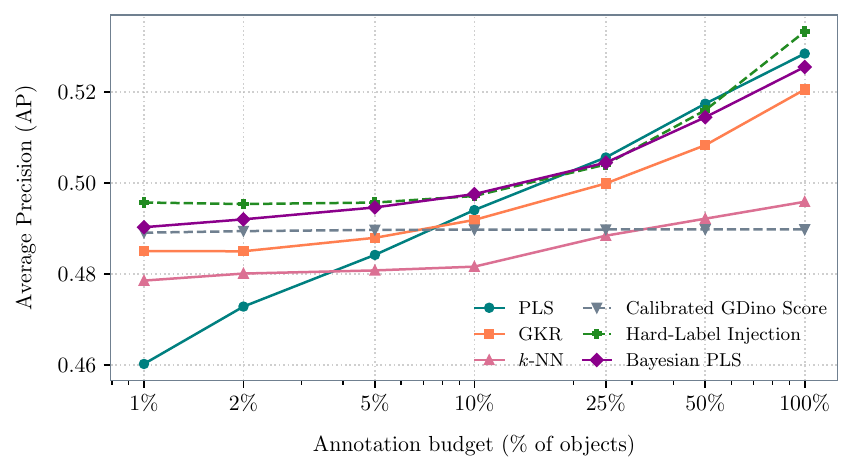}
    \caption{Average Precision (AP\textsubscript{50:95})~\cite{COCO} on COCO 2017 validation data with soft labels~\cite{benchmark} depending on annotation budget for our propagation method (PLS) against Gaussian Kernel Regression, $k$-NN regression, calibrated Grounding DINO confidence, and Hard-Label Injection. We also include a Bayesian approach of our method based on the initial Grounding DINO score. Here, the goal is essentially to resort predictions based on obtained labels such that AP increases.}
    \label{fig:AP_comparison}
\end{figure}

We measure COCO annotation-quality in both AS and AP. For these metrics, PLS propagates two independent signals jointly over one graph. Since AS focuses on soft labels, it uses the propagated label information directly, i.e., each evaluate detection's label is propagated resulting in soft label estimates for each object. In contrast, AP focuses on ranking of predictions, which is why in that case, we propagate the binary TP/FP signal obtained by IoU-matching each detection against the soft ground truth. Hence, label propagation re-scores the original predictions based on the label information available. This can be interpreted as metaclassification, i.e., learning which predictions were correct and which ones were incorrect by propagating received label information over the similarity graph. AP is then computed with the standard COCO evaluation protocol~\cite{COCO} against the soft ground truth. To derive hard labels from soft labels, they are thresholded such that all objects with a minimal class probability of at least 0.5 for any non-background class are included in the set of annotations as proposed by~\cite{benchmark}.

\begin{figure}[tb]
    \centering
    \begin{subfigure}[t]{0.48\linewidth}
        \includegraphics[width=\linewidth]{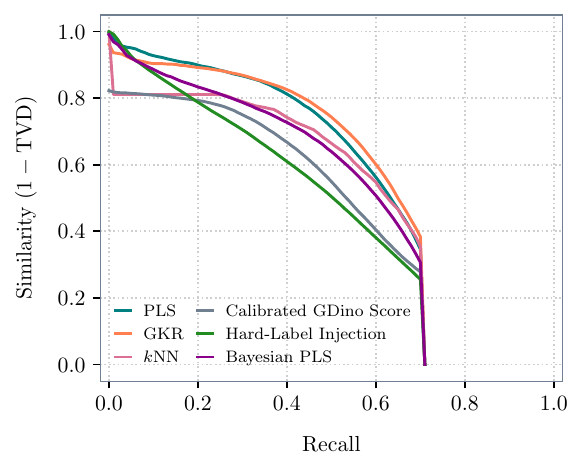}
        \caption{}
    \label{subfig:as_curves_iou_50}
    \end{subfigure}
    \hfill    
    \begin{subfigure}[t]{0.48\linewidth}
        \includegraphics[width=\linewidth]{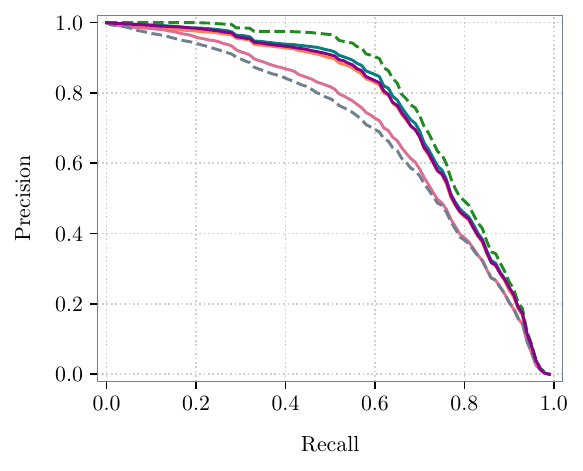}
        \caption{}
    \label{subfig:ap_curves_iou_50}
    \end{subfigure}
    
    \caption{Similarity-Recall Curve (a) and Precision-Recall Curve (b) on COCO for an IoU threshold of 0.5. To determine similarity in terms of 1-TVD and precision, estimated and true soft labels are compared, where predictions are sorted by confidence as usual in COCO-style evaluation. The displayed curves are the average of all 80 class-specific curves.}
    \label{fig:as_and_ap_curves_iou_50}
\end{figure}
\begin{figure}[tb]
    \centering
    \begin{subfigure}[t]{0.48\linewidth}
        \includegraphics[width=\linewidth]{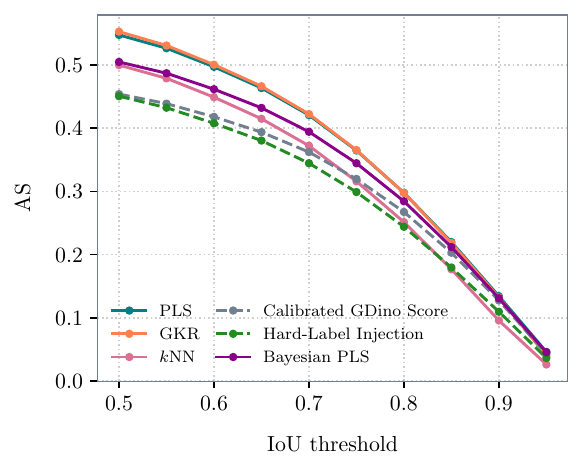}
        \caption{}
    \end{subfigure}
    \hfill    
    \begin{subfigure}[t]{0.48\linewidth}
        \includegraphics[width=\linewidth]{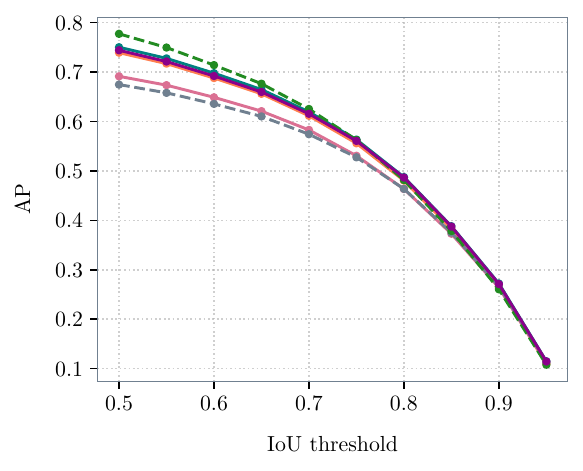}
        \caption{}
    \end{subfigure}
    \caption{Average Similarity (a) and Average Precision (b) depending on the IoU threshold used for GT matching of predictions. A common pattern is that improvements over the Grounding DINO baseline through label propagation are most prominent for lower IoU thresholds.}
    \label{fig:as_and_ap_depending_on_iou}
\end{figure}

% Other datasets

\begin{figure*}[tb]
    \centering
    \begin{subfigure}[t]{0.32\linewidth}
        \includegraphics[width=\linewidth]{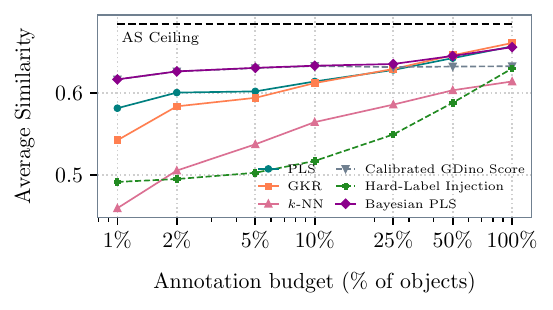}
        \caption{Pascal VOC}
    \end{subfigure}
    \hfill    
    \begin{subfigure}[t]{0.32\linewidth}
        \includegraphics[width=\linewidth]{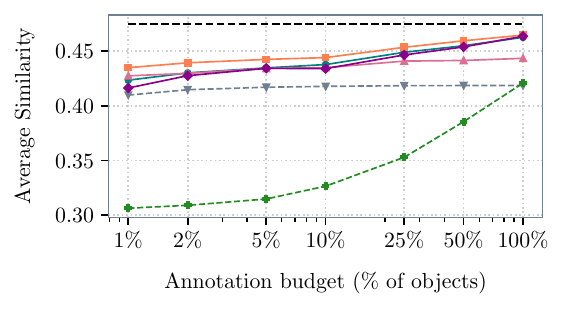}
        \caption{KITTI}
    \end{subfigure}
    \hfill    
    \begin{subfigure}[t]{0.32\linewidth}
        \includegraphics[width=\linewidth]{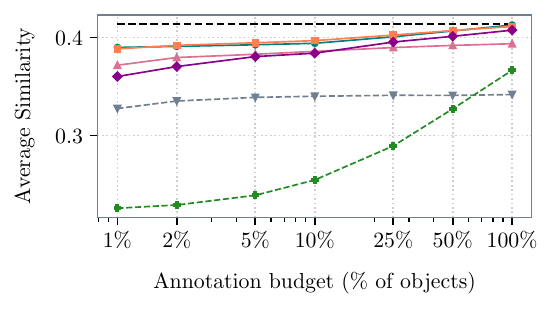}
        \caption{Cityscapes}
    \end{subfigure}
    
    \caption{Performance in terms of Average Similarity depending on the provided annotation budget for three datasets. Hyperparameters were swept through a grid search  and optimal values for each annotation budget are displayed.}
    \label{fig:AS_other_datasets}
\end{figure*}

\begin{figure*}[tb]
    \centering
    \begin{subfigure}[t]{0.32\linewidth}
        \includegraphics[width=\linewidth]{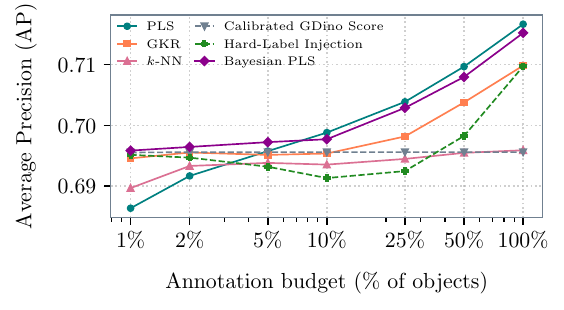}
        \caption{Pascal VOC}
    \end{subfigure}
    \hfill    
    \begin{subfigure}[t]{0.32\linewidth}
        \includegraphics[width=\linewidth]{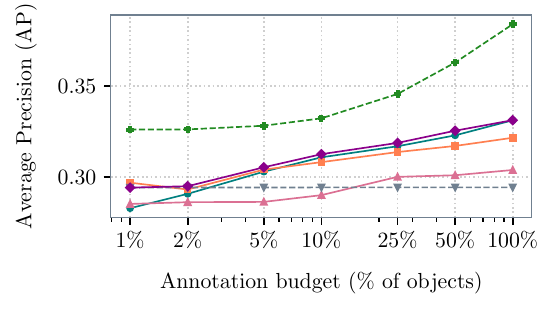}
        \caption{KITTI}
    \end{subfigure}
    \hfill    
    \begin{subfigure}[t]{0.32\linewidth}
        \includegraphics[width=\linewidth]{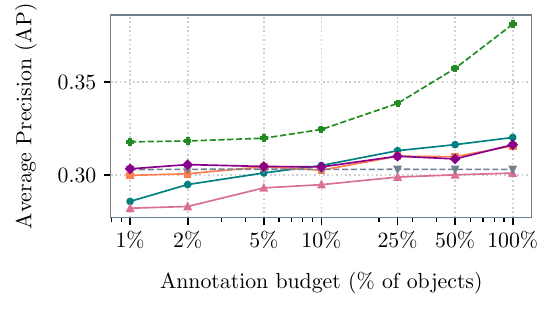}
        \caption{Cityscapes}
    \end{subfigure}
    
    \caption{Performance in terms of Average Precision depending on the provided annotation budget for three datasets. Hyperparameters were swept through a grid search  and optimal values for each annotation budget are displayed.}
    \label{fig:AP_other_datasets}
\end{figure*}

\paragraph{AP Results.}
Performance in terms of AP on COCO is displayed in~\autoref{fig:AP_comparison}. The calibrated Grounding DINO baseline is budget-invariant since recalibrating raw confidence scores does not change their ranking with the isotonic regression model applied. It serves as a fixed reference point and does not benefit from more data. PLS alone starts below this reference at low budgets and overtakes it with a budget of $5$--$10\%$, then improves further with more supervision. Hence, graph-based label spreading needs a critical mass of confirmed TPs and FPs in the graph before its re-ranking becomes competitive with the raw detector score. Bayesian blending of the PLS estimate with the calibrated Grounding DINO prior removes this weakness entirely. The blend matches or exceeds the baseline at every budget. The Bayesian PLS is closely trailing the ranking based on hard label injection, which is even briefly surpassed by pure PLS. Hard label injection puts confirmed TPs at the top of the confidence ranking and confirmed FPs at the very bottom while preserving original detector ordering between TPs and FPs (we assign TPs a score of $1 - 10^{-6}(1-\hat{s})$, and FPs a score of $10^{-6}\hat{s}$ where $\hat{s}$ is the original score). Absolute AP gains of Bayesian PLS over the Grounding DINO baseline grow from roughly 1 percentage point at 10\% budget (49.8 vs.\ 48.9) to about 4 percentage points at 100\% (52.6 vs.\ 48.9).

\paragraph{AP and AS Curves.}
We take a closer look on the AS and AP curves of each method at a fixed IoU threshold of 0.5 on COCO. Looking at the Average Similarity Curves in \autoref{subfig:as_curves_iou_50}, we observe that alignment with soft labels is significantly improved for propagation-based methods, compared to the Grounding DINO calibration baseline and Hard-Label injection alone. PLS seems to achieve best calibration of soft labels for most confident cases while GKR catches up towards the lower-confidence regime. We used the 10\% budget scenario to generate the AS curves. The AP curves at a budget of 100\% in \autoref{subfig:ap_curves_iou_50} show that the optimal sorting of Hard-Label injection is not reached by the propagation-based methods but they significantly improve the Grounding DINO baseline. Further, we investigate AS and AP at different IoU thresholds in \autoref{fig:as_and_ap_depending_on_iou} and notice that improvements over the Grounding DINO baseline are most prominent for low IoU thresholds, while the Grounding DINO score alone cannot be improved significantly in terms of AS and AP for very well-localized objects. Hence, localization quality seems to correlate with calibration of detection confidence.

%% ---- AS/AP table (two subtables) ----
\begin{table*}[t]
\centering
\caption{Overview of performance in terms of AS and AP (\%) across datasets, methods, and three annotation budgets. Best values per budget is highlighted in bold, second best is underlined.}
\label{tab:as_ap_comparison}
\begin{subtable}[t]{\linewidth}
\centering
\caption{Average Similarity (AS)}
\label{tab:as_comparison}
\resizebox{0.95\linewidth}{!}{%
\begin{tabular}{l|rrr|rrr|rrr|rrr}
\toprule
 & \multicolumn{3}{c}{COCO} & \multicolumn{3}{c}{Cityscapes} & \multicolumn{3}{c}{KITTI} & \multicolumn{3}{c}{Pascal VOC} \\
Method & 1\% & 10\% & 100\% & 1\% & 10\% & 100\% & 1\% & 10\% & 100\% & 1\% & 10\% & 100\% \\
\midrule
PLS & \textbf{28.5} & \underline{35.2} & \underline{40.6} & \textbf{39.0} & \underline{39.4} & \textbf{41.3} & 42.3 & \underline{43.8} & 46.3 & 58.2 & \underline{61.4} & \underline{65.7} \\
Bayesian PLS & \underline{27.5} & 34.0 & 39.8 & 36.0 & 38.4 & 40.7 & 41.6 & 43.4 & \underline{46.4} & \textbf{61.7} & \textbf{63.3} & 65.6 \\
GKR & 26.6 & \textbf{35.3} & \textbf{41.6} & \underline{38.8} & \textbf{39.7} & \underline{41.1} & \textbf{43.5} & \textbf{44.4} & \textbf{46.5} & 54.3 & 61.2 & \textbf{66.1} \\
$k$-NN & 23.8 & 30.8 & 36.0 & 37.2 & 38.6 & 39.4 & \underline{42.7} & 43.5 & 44.3 & 46.0 & 56.5 & 61.4 \\
Calibrated GDino Score & 26.4 & 30.3 & 30.7 & 32.8 & 34.0 & 34.2 & 41.0 & 41.8 & 41.9 & \underline{61.6} & \textbf{63.3} & 63.3 \\
Hard-Label Injection & 26.9 & 28.9 & 38.3 & 22.6 & 25.5 & 36.7 & 30.6 & 32.6 & 42.1 & 49.2 & 51.7 & 63.0 \\
\bottomrule
\end{tabular}
}
\end{subtable}
\bigskip
\vspace{0.01cm}
\begin{subtable}[t]{\linewidth}
\centering
\caption{Average Precision (AP)}
\label{tab:ap_comparison}
\resizebox{0.95\linewidth}{!}{%
\begin{tabular}{l|rrr|rrr|rrr|rrr}
\toprule
 & \multicolumn{3}{c}{COCO} & \multicolumn{3}{c}{Cityscapes} & \multicolumn{3}{c}{KITTI} & \multicolumn{3}{c}{Pascal VOC} \\
Method & 1\% & 10\% & 100\% & 1\% & 10\% & 100\% & 1\% & 10\% & 100\% & 1\% & 10\% & 100\% \\
\midrule
PLS & 46.0 & 49.4 & \underline{52.9} & 28.6 & \underline{30.5} & \underline{32.0} & 28.3 & 31.1 & \underline{33.1} & 68.6 & \textbf{69.9} & \textbf{71.7} \\
Bayesian PLS & \underline{49.0} & \textbf{49.8} & 52.6 & \underline{30.3} & 30.4 & 31.6 & 29.4 & \underline{31.2} & \underline{33.1} & \textbf{69.6} & \underline{69.8} & \underline{71.5} \\
GKR & 48.5 & 49.2 & 52.1 & 30.0 & 30.3 & 31.6 & \underline{29.7} & 30.8 & 32.2 & \underline{69.5} & 69.5 & 71.0 \\
$k$-NN & 47.9 & 48.2 & 49.6 & 28.2 & 29.5 & 30.1 & 28.5 & 29.0 & 30.4 & 69.0 & 69.4 & 69.6 \\
Calibrated GDino Score & 48.9 & 49.0 & 49.0 & \underline{30.3} & 30.3 & 30.3 & 29.4 & 29.4 & 29.4 & \textbf{69.6} & 69.6 & 69.6 \\
Hard-Label Injection & \textbf{49.6} & \underline{49.7} & \textbf{53.3} & \textbf{31.8} & \textbf{32.4} & \textbf{38.1} & \textbf{32.6} & \textbf{33.2} & \textbf{38.4} & \underline{69.5} & 69.1 & 71.0 \\
\bottomrule
\end{tabular}
}
\end{subtable}
\end{table*}

\paragraph{Performance on other Datasets.}
We displayed Average Similarity (AS) as a function of annotation budget for the COCO dataset~\cite{COCO} in the main manuscript. Here, we additionally discuss performance in terms of Average Precision (AP) on COCO as well as results on other datasets. COCO AP results are displayed in~\autoref{fig:AP_comparison}, AS and AP results for Pascal VOC~\cite{PascalVOC}, KITTI~\cite{kitti}, and Cityscapes~\cite{Cityscapes} are displayed in figures \ref{fig:AS_other_datasets} and \ref{fig:AP_other_datasets} while \autoref{tab:as_ap_comparison} gives an overview of results.

On Cityscapes and KITTI, results are similar to COCO for AS but quite different for AP. Propagation-based methods reach the AS ceiling almost immediately while calibrated Grounding DINO and Hard-Label Injection trail these methods by a wide margin on AS. This indicates that label propagation is very beneficial to approximate the soft labels on these two datasets, compared to post-hoc calibration alone.
In terms of AP, however, Hard-Label Injection performs significantly better than propagation-based methods. This disagreement of AS and AP reflects that propagation-based methods are well suited for approximating the class distribution of objects, while the direct use of GT leads to the best ranking of detections that graph-based label spreading and kernel regression cannot fully match through the diffusion of binary TP/FP information.

Pascal VOC results are slightly different again. Here, calibrated Grounding DINO scores are
already a strong baseline at minimal budget. Because of this strong
starting point, propagation-based methods only overtake baselines in both AS and AP once the annotation budget grows to roughly $25\%$.
VOC is also the only dataset where Hard-Label
Injection's AP is non-monotonic in budget, dipping from $69.5\%$ at $1\%$ to
$69.1\%$ at $10\%$ before recovering to $71.0\%$ at $100\%$. One reason for this may be that Hard-Label
Injection re-ranks based on a matching computed up front, so a duplicate detection for an already-matched GT object can occasionally be
sampled and confirmed as a TP instead of the genuine true-positive box, which may have a better localization.
This steal of the GT bounding box then injects a false positive near the top of the ranking. Notably, at $100\%$ budget, both PLS and Bayesian
PLS ($71.7\%$ and $71.5\%$ AP respectively) surpass Hard-Label Injection ($71.0\%$ AP). So smoothing label information across the graph improves on the oracle ranking at an IoU threshold of $0.5$, which is the optimal ranking at this threshold but potentially not at higher thresholds.

\appendixheading{Effect of Hyperparameters}
\label{sec: Appendix - Hyperparameter Study}

\begin{figure*}[tb]
    \centering
    \begin{subfigure}[t]{0.32\linewidth}
        \includegraphics[width=\linewidth]{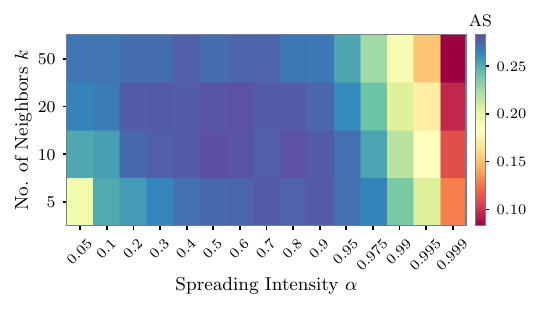}
        \caption{1\% Budget}
    \end{subfigure}
    \hfill    
    \begin{subfigure}[t]{0.32\linewidth}
        \includegraphics[width=\linewidth]{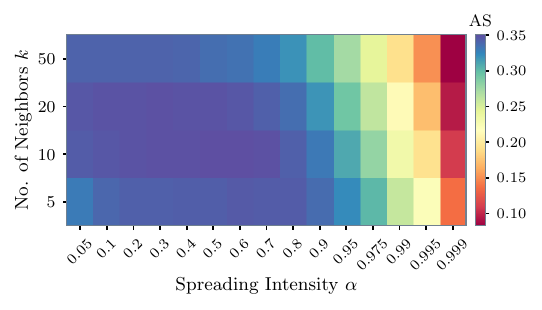}
        \caption{10\% Budget}
    \end{subfigure}
    \hfill    
    \begin{subfigure}[t]{0.32\linewidth}
        \includegraphics[width=\linewidth]{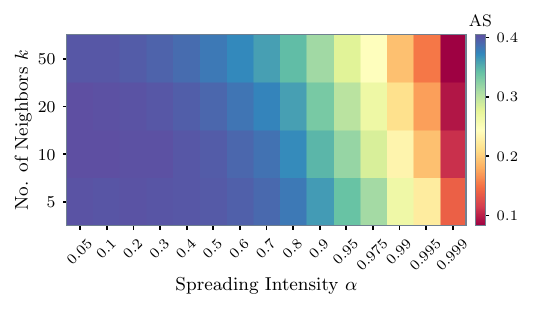}
        \caption{100\% Budget}
    \end{subfigure}
    
    \caption{Hyperparameter analysis of our method for both spreading intensity $\alpha$ and the number of neighbors $k$. We evaluate Average Similarity on COCO 2017 validation data with soft labels~\cite{benchmark} for each parameter combination.}
    \label{fig:hyperparameter_heatmaps}
\end{figure*}

\begin{figure*}[tb]
    \centering
    \begin{subfigure}[t]{0.32\linewidth}
        \includegraphics[width=\linewidth]{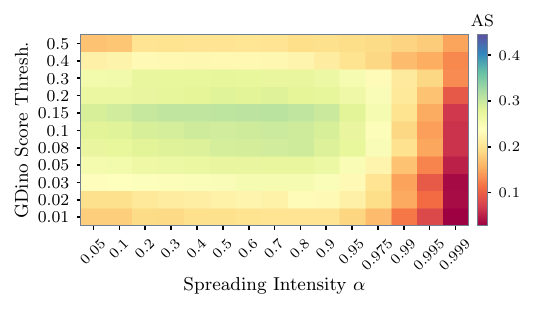}
        \caption{1\% Budget}
    \end{subfigure}
    \hfill    
    \begin{subfigure}[t]{0.32\linewidth}
        \includegraphics[width=\linewidth]{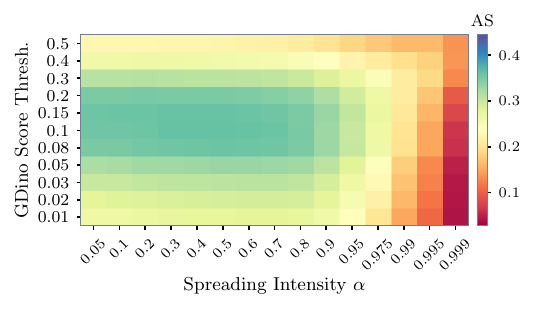}
        \caption{10\% Budget}
    \end{subfigure}
    \hfill    
    \begin{subfigure}[t]{0.32\linewidth}
        \includegraphics[width=\linewidth]{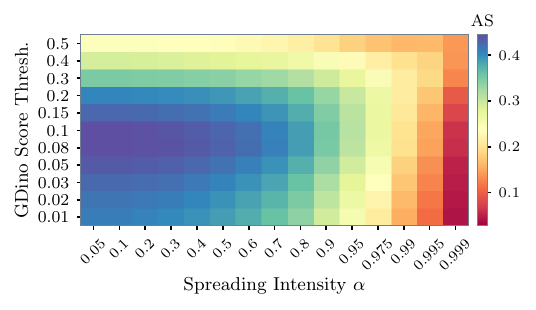}
        \caption{100\% Budget}
    \end{subfigure}
    
    \caption{Hyperparameter analysis of our method for the Grounding DINO score threshold applied as well as the spreading intensity $\alpha$ for a fixed number of neighbors $k=20$. We evaluate Average Similarity on COCO 2017 validation data with soft labels~\cite{benchmark} for each parameter combination.}
    \label{fig:grounding_dino_score_heatmaps}
\end{figure*}

To assess the sensitivity of our method to hyperparameters, we sweep the number of neighbors $k$ and the spreading intensity $\alpha$ for three annotation budgets. We evaluate performance in terms of achieved Average Similarity (AS) on soft labeled COCO data as discussed in the main paper and display the results in~\autoref{fig:hyperparameter_heatmaps}. 
Overall, AS is relatively robust to both parameters, varying only mildly across most of the grid while degrading for extremely large values of $\alpha$. The dependency on hyperparameters is budget-conditional. At small budgets, where few labels are available, heavier propagation is preferred, and the heatmap shows poor performance in the regime of low $k$ and low $\alpha$. In this case, information is presumably not spread sufficiently, resulting in sparsely-labeled regions. At larger budgets, labeled nodes are already dense enough that little propagation is needed, and the optimum shifts to small $k$ and small $\alpha$, which avoids oversmoothing.

% Grounding DINO score
To assess how the confidence threshold applied to raw Grounding DINO detections for pre-filtering affects performance, we swept it jointly with the spreading intensity $\alpha$ (score threshold $t\in[0.01, 0.5]$, $\alpha\in[0.05, 0.999]$, $k=20$ fixed) across three annotation budgets, the results are shown in~\autoref{fig:grounding_dino_score_heatmaps}.
The score threshold substantially affects AS, where lower thresholds lead to higher scores in this recall-sensitive metric across every budget and every $\alpha$. At a lower threshold, more detections are retained, which expands the object graph by almost two orders of magnitude (from $\approx 20,000$ at $t=0.5$ to well over a million nodes at $t=0.01$). The role of $\alpha$ is threshold-dependent. At low thresholds, where the graph is large and densely connected, AS does not vary significantly with $\alpha$ and reduces only marginally as $\alpha \to 1$. At high thresholds, where only the most confident detections form the graph, AS increases from low $\alpha$ and plateaus at around $\alpha \approx 0.2$–$0.6$ before falling off as $\alpha$ approaches 1. This is consistent across all three budgets, and matches the intuition that a smaller graph offers fewer similar neighbors to spread over. Choosing $\alpha$ too close to 1 leads to over-smoothing in that case. Overall, a large $\alpha$ can be chosen if a small score threshold is applied such that the underlying graph is sufficiently large whereas at higher thresholds, a moderate $\alpha$ is preferable.

\appendixheading{Active Querying}
\label{sec:appendix-active_querying}

\begin{figure}[tb]
    \centering
    \begin{subfigure}[t]{0.48\linewidth}
        \includegraphics[width=\linewidth]{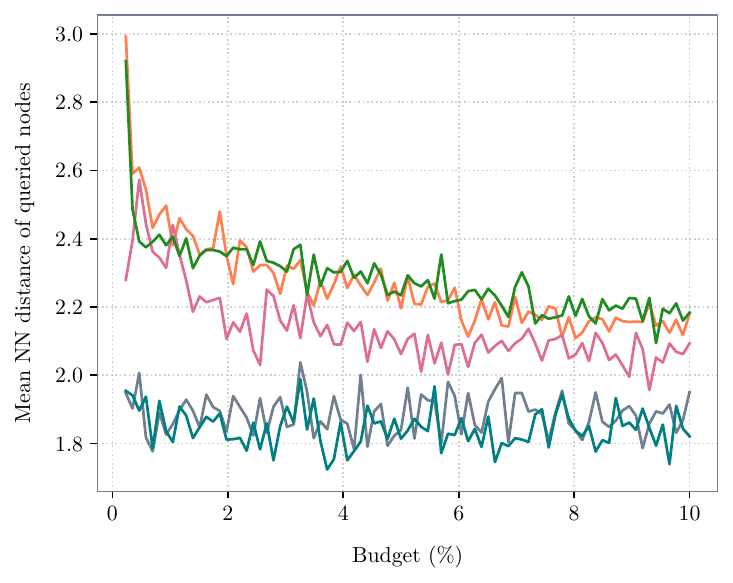}
        \caption{}
    \end{subfigure}
    \hfill    
    \begin{subfigure}[t]{0.48\linewidth}
        \includegraphics[width=\linewidth]{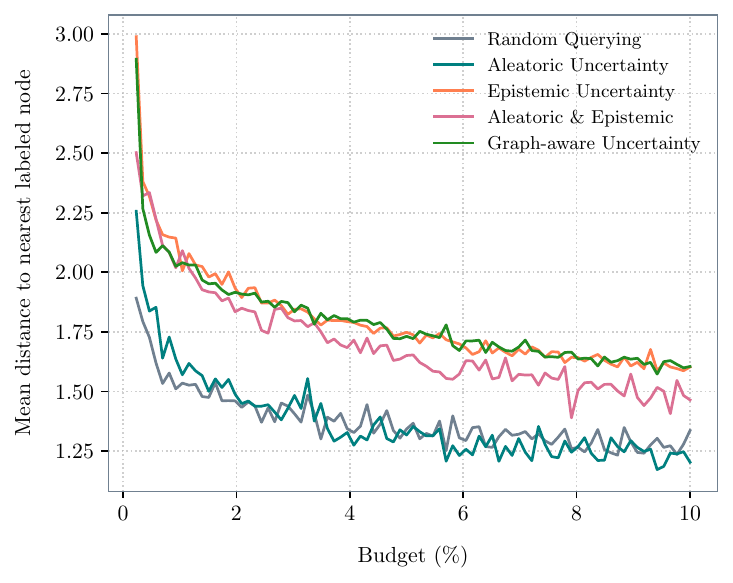}
        \caption{}
    \end{subfigure}
    \caption{Coverage of labeled nodes during active querying in terms of two metrics where higher indicates better spread of labels across the feature space. In (a),  the average $k$-NN distance between labeled nodes is displayed as a moving average with a window size of 100. Similarly, the redundancy of labeled nodes is captured in (b) by the average distance to any labeled node for all nodes in the graph.}
    \label{fig:AL_metrics}
\end{figure}

In terms of active querying, we evaluate five different strategies in terms of AS and AP in the main paper.
Here, we give additional details on this experiment and provide more results that give an intuitive explanation of the strategies results.

For this study, we consider the COCO dataset, similar to the main data quality evaluation, we construct the graph on top of Grounding DINO embeddings, thresholded at a minimal confidence of 0.2.
We use a fix parameter combination of $k=20$ neighbors and $\alpha = 0.9$ for PLS and query one label at a time for each strategy. 
Based on the current set of labels, PLS with the fixed parameters is applied to estimate soft labels and objectness scores. 
We evaluate performance in terms of AS and AP every 100 queries as this is the most compute heavy part of the experiment. 
The random strategy queries a random node for labeling at a time, drawn uniformly from the set of all features in the graph. 
Epistemic uncertainty querying leverages the normalization we apply to propagation scores such that a labeled node gets a propagation score of 1 and neighboring nodes get partial feedback. Hence, cumulative propagation scores can be interpreted as the number of virtual labels obtained. The node to label next is determined by the cumulative label feedback obtained so far, i.e., the node with minimal received label information (summed over all classes) is queried next.  
The aleatoric uncertainty strategy focuses on estimated soft labels directly and queries the next label for the node whose soft label estimate has the highest ambiguity in terms of Shannon entropy. Shannon entropy of an estimated soft label $\hat{p_j}\in \Delta^{C+1}$ is given as
\begin{equation}
    \mathcal{H}(\hat{p}_j) = -\sum_{i=1}^{C+1} \hat{p}_{j,i} \log_2(\hat{p}_{j,i}),
\end{equation}
where the convention is to set a summand to zero whenever an appearing probability is zero.
For the mixture strategy, we first normalize scores such that the node with the highest score is scored with 1 and the lowest with 0 through the corresponding affine transformation of scores. We then weight both uncertainty scores equally and query the node with the highest combined score. 
For the Graph-aware strategy, we rely on the Wilson score intervals~\cite{Wilson1927} that can be derived from the virtual number of labels received per class in each node. Specifying a confidence level of $95\%$, this gives an interval width for each estimated class probability in the soft label. We sum the individual widths over all classes to obtain an aggregated width $w_i$ for node $i$ that reflects both aleatoric and epistemic uncertainty. Finally, we leverage the heat kernel to let this quantity diffuse over the graph just as in regular label spreading. For a node $j$, we obtain the score
\begin{equation*}
        s_j = \sum_{i=1}^{n}(I-\alpha S)^{-1}_{ij}\, w_i \Bigm/ \sum_{i=1}^{n} (I-\alpha S)^{-1}_{ij}.
\end{equation*}
Here, we found that the normalization with the propagated mass is necessary such that the differences in confidence interval widths $w_i$ are reflected and the resulting score $s_j$ is not depending on the graph structure only. We query the node that maximizes $s_j$ which is the node whose neighborhood exhibits the highest uncertainty in terms of Wilson score intervals.
In \autoref{fig:AL_metrics}, we see that  querying by epistemic uncertainty and the graph-based approach lead to better spread of labels compared to random querying whereas this is not the case for aleatoric uncertainty alone.

\begin{figure*}[tb]
    \centering
    \begin{subfigure}[t]{0.18\linewidth}
        \includegraphics[width=\linewidth]{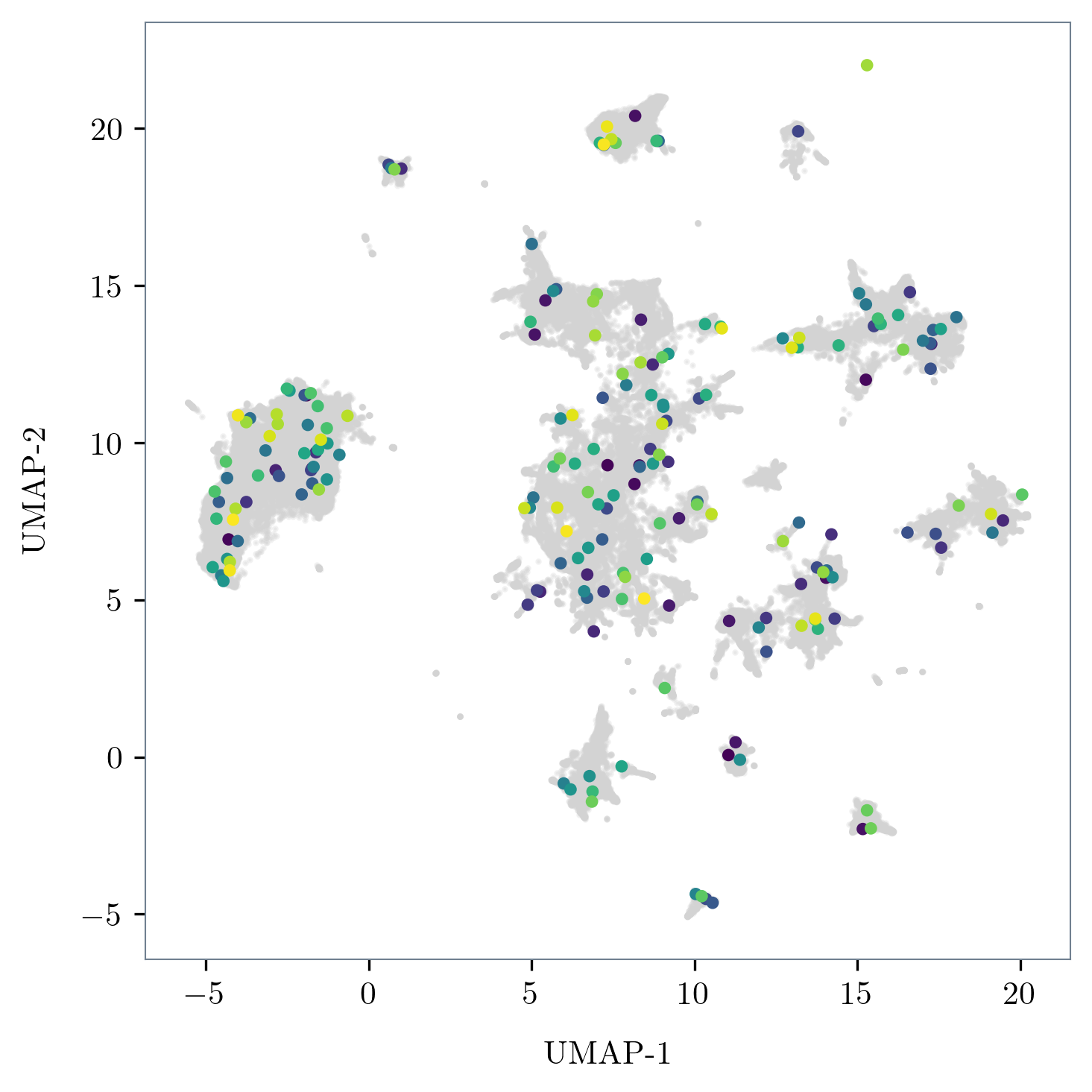}
        \caption{\footnotesize Random Querying}
    \end{subfigure}
    \hfill    
    \begin{subfigure}[t]{0.18\linewidth}
        \includegraphics[width=\linewidth]{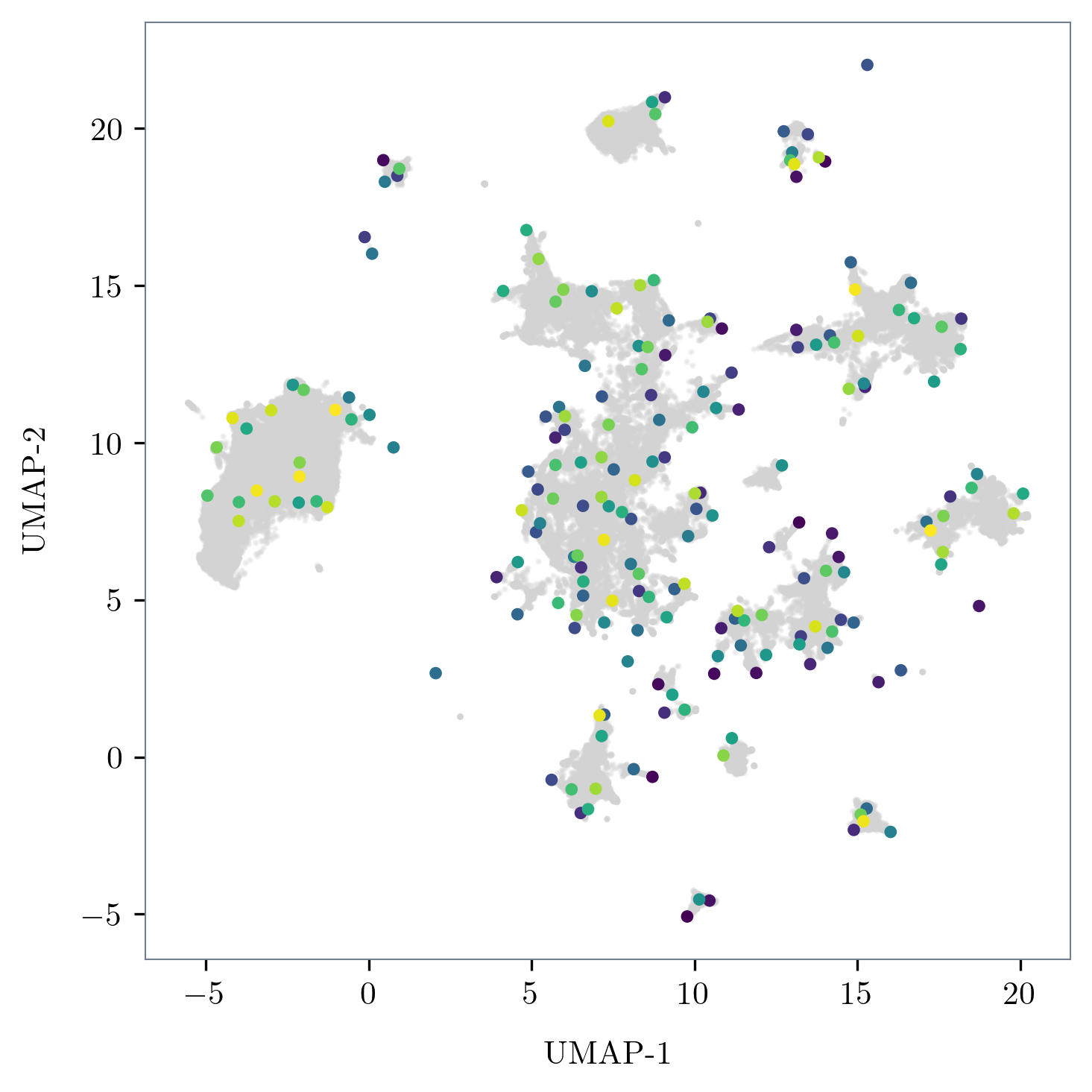}
        \caption{\footnotesize Epistemic Uncertainty}
    \end{subfigure}
    \hfill   
    \begin{subfigure}[t]{0.18\linewidth}
        \includegraphics[width=\linewidth]{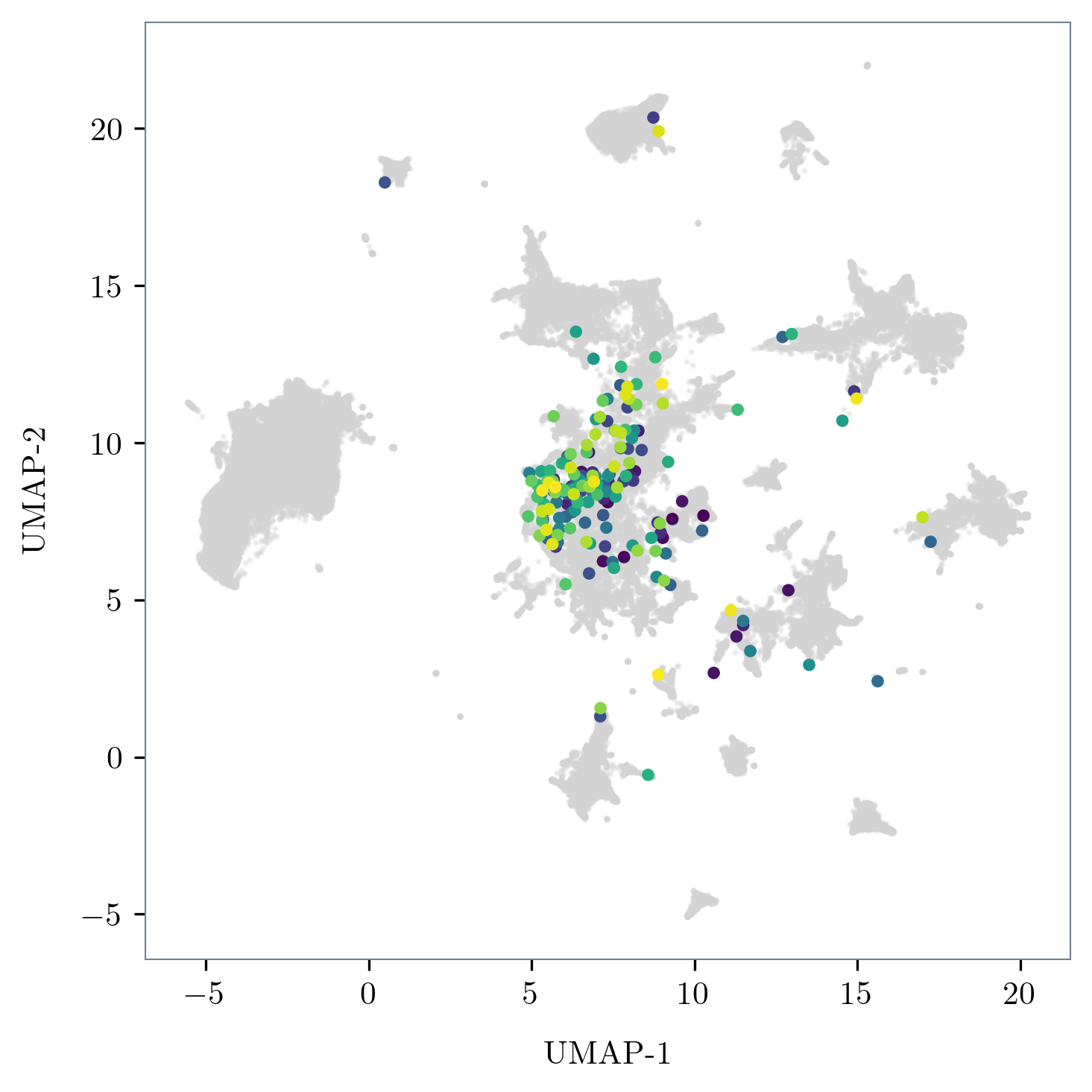}
        \caption{\footnotesize Aleatoric Uncertainty}
    \end{subfigure}
    \hfill   
    \begin{subfigure}[t]{0.18\linewidth}
        \includegraphics[width=\linewidth]{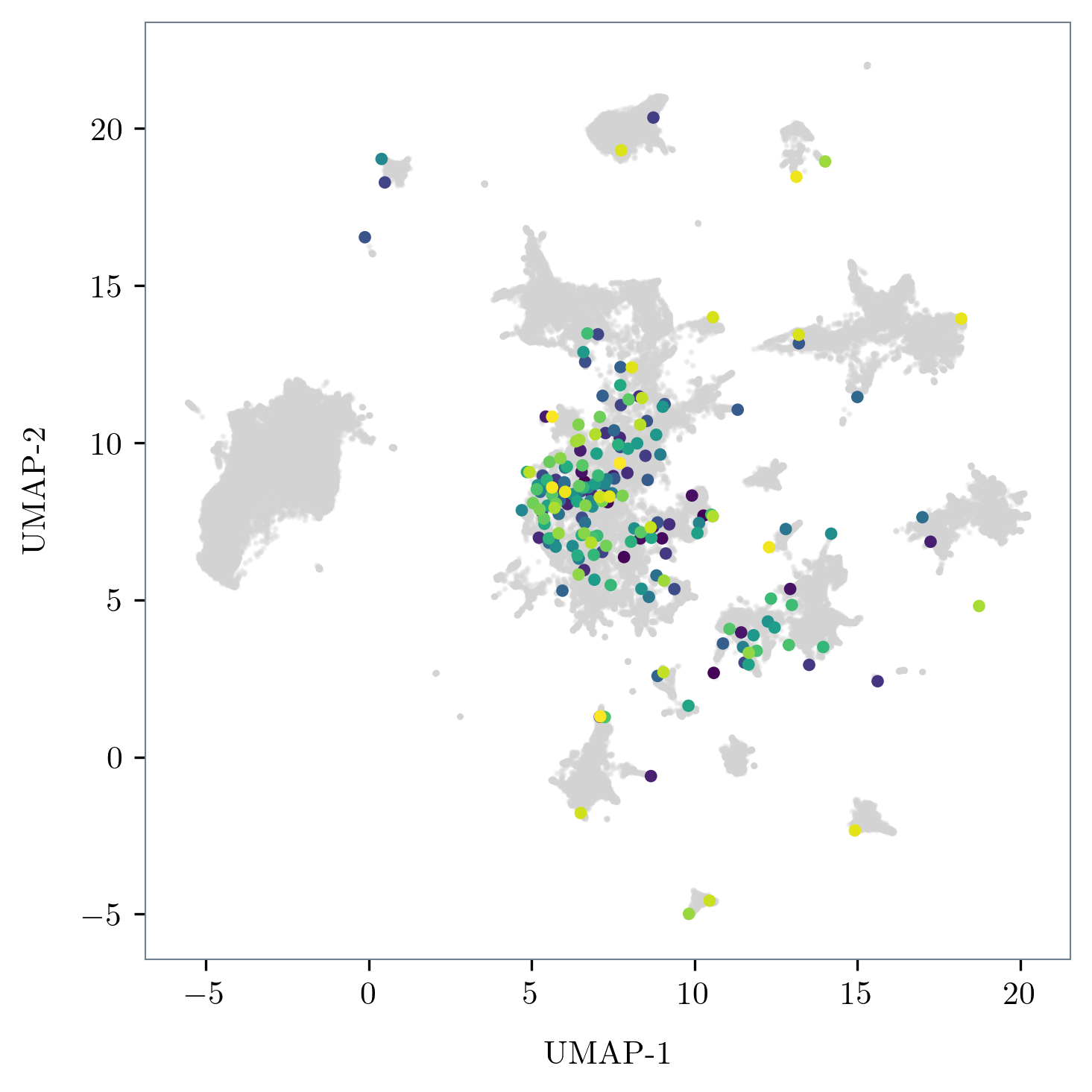}
        \caption{\footnotesize Epistemic \& Aleatoric}
    \end{subfigure}
    \hfill   
    \begin{subfigure}[t]{0.18\linewidth}
        \includegraphics[width=\linewidth]{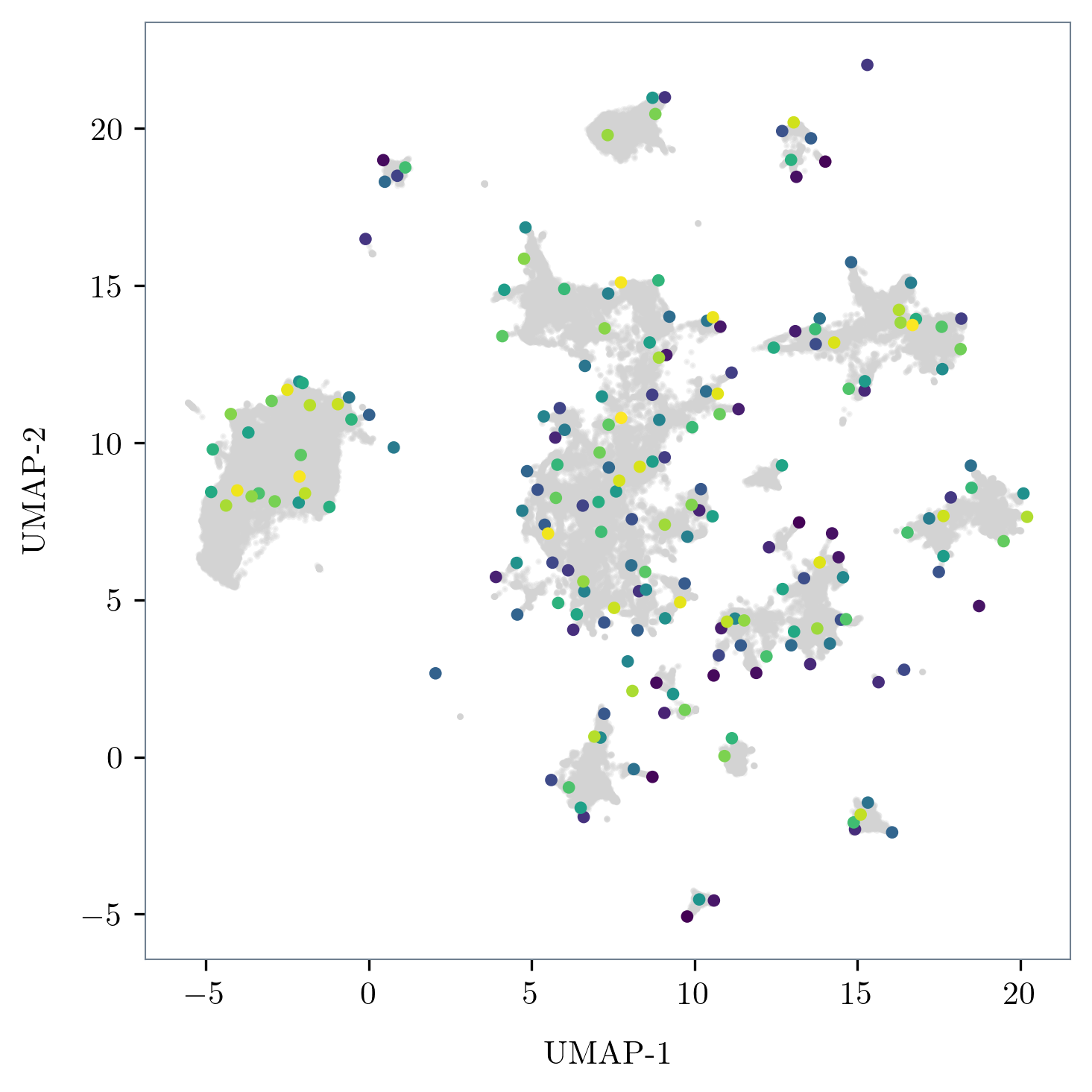}
        \caption{\footnotesize Graph-aware}
    \end{subfigure}
    \hfill   
    \begin{subfigure}[t]{0.06\linewidth}
        \includegraphics[width=\linewidth]{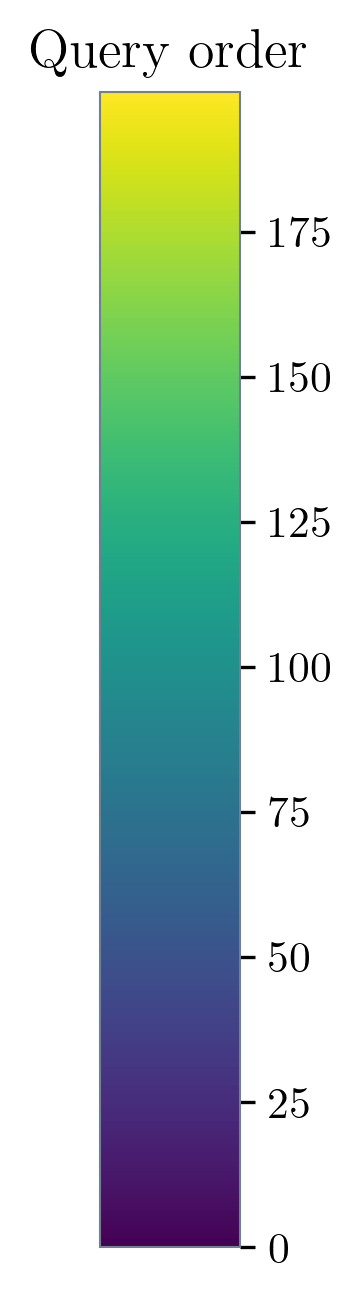}
    \end{subfigure}
    \hfill  
    \caption{First 200 Queried nodes in two-dimensional UMAP embedding space of Grounding DINO features of COCO validation data for different querying strategies.}
    \label{fig:AL_Queried_points_in_UMAP}
\end{figure*}

\appendixheading{Annotation Costs}
\label{sec: Appendix - Annotation Costs}

Here, we compare the annotation costs that can be expected using our approach for the semi-automated, uncertainty-aware annotation of object detection datasets. The results here are based on~\cite{benchmark} where annotation costs are reported for the fully manual dataset construction but also for the evaluation of individual objects by human annotators in a crowdsourcing scenario. With this detailed information on how long it takes an annotator to evaluate an object instance regarding its class affiliation, we can determine an estimate for the annotation costs associated with our approach. In particular, we use the median across all answer times per dataset as the expected annotation costs per object instance. This number is then multiplied with the number of objects to evaluate to derive the total annotation costs as listed in \autoref{tab:annotation_cost}. Comparing these numbers with the annotation costs of the fully manual approach, we conclude that the semi-automated approach is significantly more efficient in terms of labeling costs. Efficiency is driven by the automated construction of bounding boxes with Grounding DINO but also by avoiding the redundant evaluation of individual object instances to derive soft labels where we rely on probabilistic label spreading instead.
\begin{table}[t]
\centering
\caption{Estimated annotation costs for our approach depending on the dataset and annotation budget. The estimate is based on each dataset's median answer time per object and is compared to the annotation costs of fully manual annotation with soft labels as done by~\cite{benchmark}.}
\label{tab:annotation_cost}
\begin{tabular}{llrrrr}
\toprule
 & Median & \multicolumn{3}{c}{Cost at budget (h)} & Fully manual \\
\cmidrule(lr){3-5}
Dataset &  answer time (s) & 1\% & 10\% & 100\% & annotation (h) \\
\midrule
COCO & 1.14 & 0.3 & 2.7 & 27.2 & 2,005.4 \\ 
Cityscapes & 0.82 & 0.2 & 1.8 & 17.7 & 992.2 \\ 
KITTI & 1.00 & 0.3 & 2.8 & 27.9 & 1,306.9 \\ 
Pascal VOC & 0.87 & 0.1 & 1.4 & 14.1 & 1,016.9 \\ 
\bottomrule
\end{tabular}
\end{table}

\newpage
\appendixheading{Visual Examples}
\label{sec: Appendix - Visual Examples}

For qualitative inspection of estimated soft labels, we visualize objects for three distinct cases. Objects where our estimate agrees with Grounding DINO on the class, objects where the two methods predict different classes, and lastly, where the soft ground truth itself is ambiguous, i.e., the highest class probability in the soft label is below 0.9. 
For the agreement cases in \autoref{fig:visual-agreement} our method consistently predicts higher confidences than Grounding DINO's calibrated confidence, resulting in better alignment with the corresponding soft labels. 
The disagreement examples in \autoref{fig:visual-disagreement} consistently show that Grounding DINO's calibrated confidence is overly leaning toward background on genuine, correctly-localized objects. E.g., the apple in a supermarket that Grounding DINO ranks below background but where label propagation recovers a high class probability by borrowing evidence from similar, higher-confidence detections in the embedding graph. 
On ambiguous ground in \autoref{fig:visual-uncertain-gt}, our estimate and Grounding DINO's calibrated confidence agree with each other on the predicted class and, in most cases, they also reproduce the ambiguity itself. In particular, they assign a reduced confidence similar to the GT annotations that build on multiple human assignments. For the bench in the train where the soft label is split evenly between bench and couch both methods predict a mixture of bench, couch and background as well. For the partially visible knife or other cutlery Grounding DINO and our method accurately express ambiguity by predicting a mixture of knife, spoon and background, similar to the soft GT. The exception is a fence-obscured truck in the ground truth (a), where both methods nonetheless relatively confidently predict the truck class while annotators did not identify it as a truck frequently.
Overall, label propagation sharpens confidence for correct predictions, corrects cases where Grounding DINO was under-confident, and preserves ambiguity in cases where human annotators are uncertain as well. A similar pattern can be observed for the examples on the other three datasets in figures~\ref{fig:visual-agreement-other}, \ref{fig:visual-disagreement-other}, and \ref{fig:visual-uncertain-gt-other}.

\begin{figure*}[p]
    \centering
    \captionsetup[subfigure]{skip=1pt}
    \begin{subfigure}[b]{0.29\textwidth}
        \includegraphics[width=\linewidth]{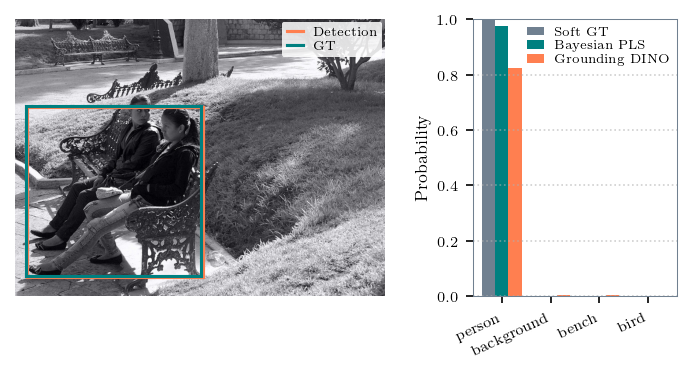}
        \caption{}
        \label{fig:visual-agreement-0}
    \end{subfigure}
    \hspace{2em}
    \begin{subfigure}[b]{0.29\textwidth}
        \includegraphics[width=\linewidth]{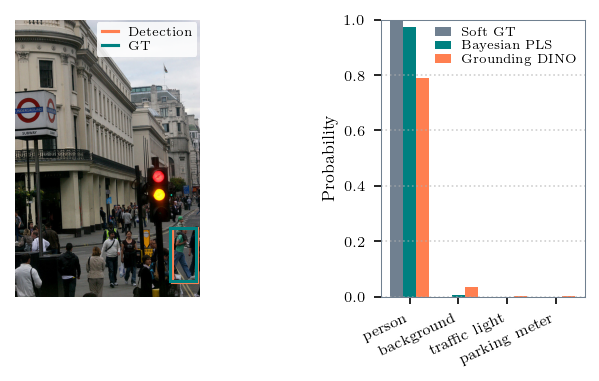}
        \caption{}
        \label{fig:visual-agreement-1}
    \end{subfigure}
    \par
    \begin{subfigure}[b]{0.29\textwidth}
        \includegraphics[width=\linewidth]{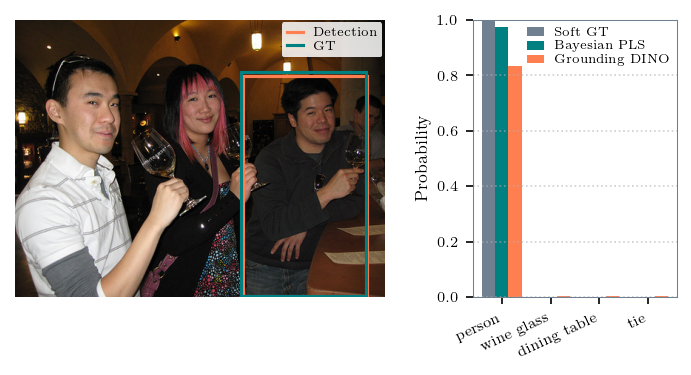}
        \caption{}
        \label{fig:visual-agreement-2}
    \end{subfigure}
    \hspace{2em}
    \begin{subfigure}[b]{0.29\textwidth}
        \includegraphics[width=\linewidth]{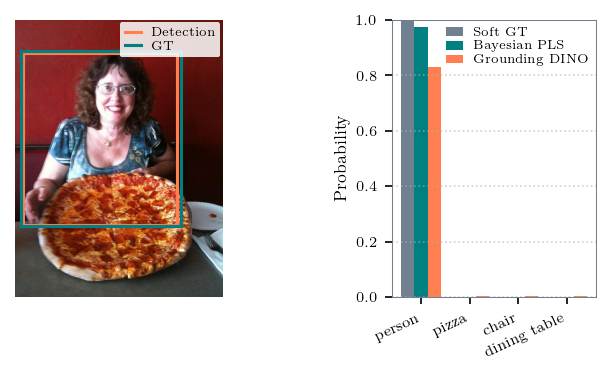}
        \caption{}
        \label{fig:visual-agreement-3}
    \end{subfigure}
    \caption{Examples where our method and Grounding DINO agree on the predicted class.}
    \label{fig:visual-agreement}
\end{figure*}

\begin{figure*}[p]
    \centering
    \captionsetup[subfigure]{skip=1pt}
    \begin{subfigure}[b]{0.29\textwidth}
        \includegraphics[width=\linewidth]{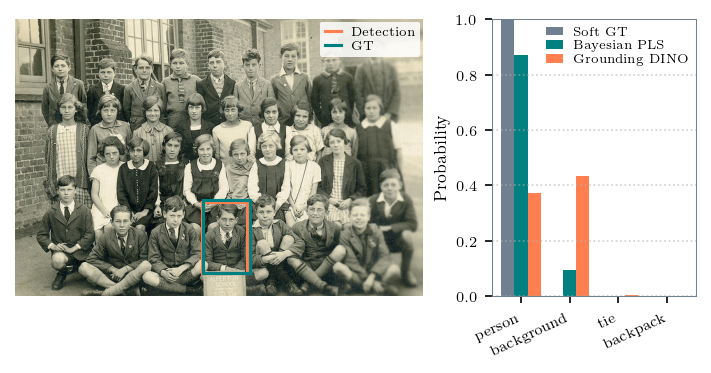}
        \caption{}
        \label{fig:visual-disagreement-0}
    \end{subfigure}
    \hspace{2em}
    \begin{subfigure}[b]{0.29\textwidth}
        \includegraphics[width=\linewidth]{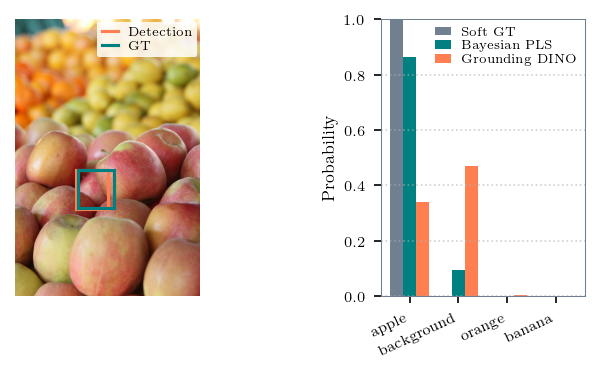}
        \caption{}
        \label{fig:visual-disagreement-1}
    \end{subfigure}
    \par
    \begin{subfigure}[b]{0.29\textwidth}
        \includegraphics[width=\linewidth]{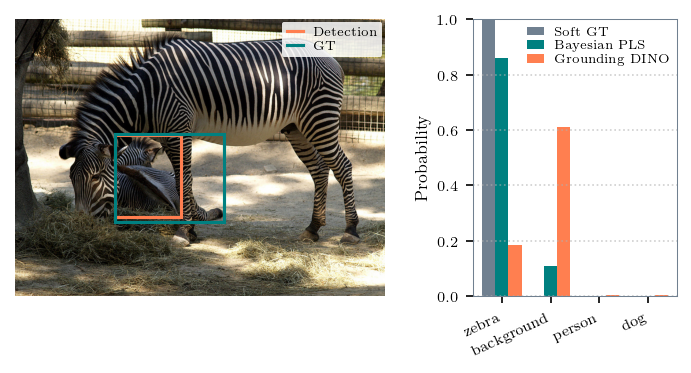}
        \caption{}
        \label{fig:visual-disagreement-2}
    \end{subfigure}
    \hspace{2em}
    \begin{subfigure}[b]{0.29\textwidth}
        \includegraphics[width=\linewidth]{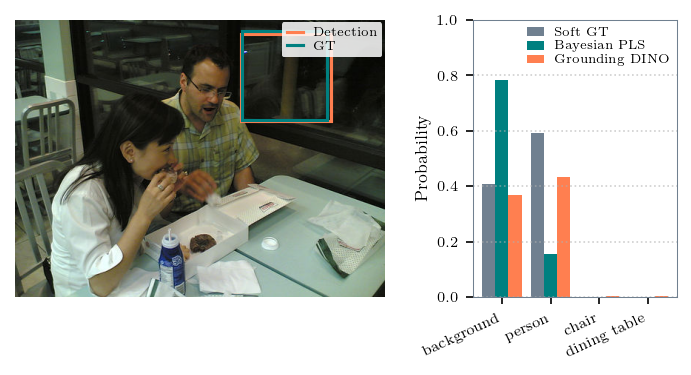}
        \caption{}
        \label{fig:visual-disagreement-3}
    \end{subfigure}
    \caption{Examples where our method and Grounding DINO disagree on the predicted class.}
    \label{fig:visual-disagreement}
\end{figure*}

\begin{figure*}[p]
    \centering
    \captionsetup[subfigure]{skip=1pt}
    \begin{subfigure}[b]{0.29\textwidth}
        \includegraphics[width=\linewidth]{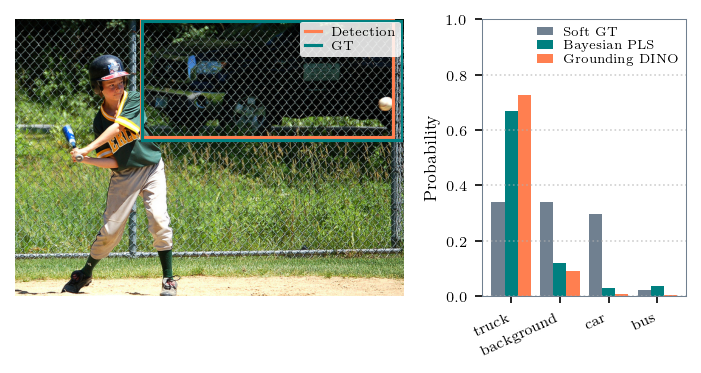}
        \caption{}
        \label{fig:visual-uncertain-gt-0}
    \end{subfigure}
    \hspace{2em}
    \begin{subfigure}[b]{0.29\textwidth}
        \includegraphics[width=\linewidth]{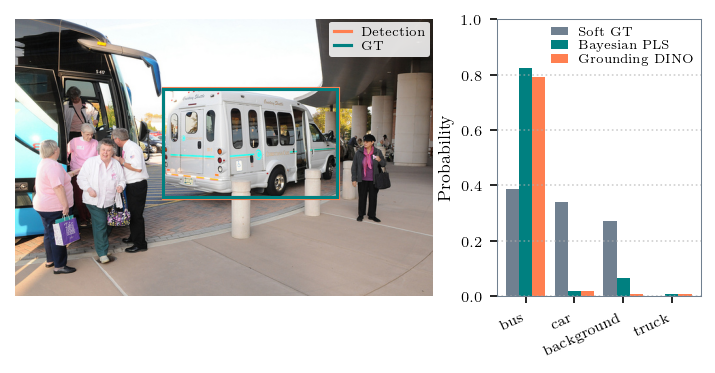}
        \caption{}
        \label{fig:visual-uncertain-gt-1}
    \end{subfigure}
    \par
    \begin{subfigure}[b]{0.29\textwidth}
        \includegraphics[width=\linewidth]{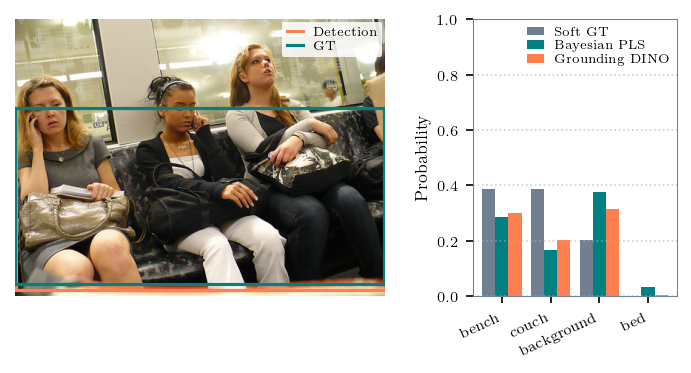}
        \caption{}
        \label{fig:visual-uncertain-gt-2}
    \end{subfigure}
    \hspace{2em}
    \begin{subfigure}[b]{0.29\textwidth}
        \includegraphics[width=\linewidth]{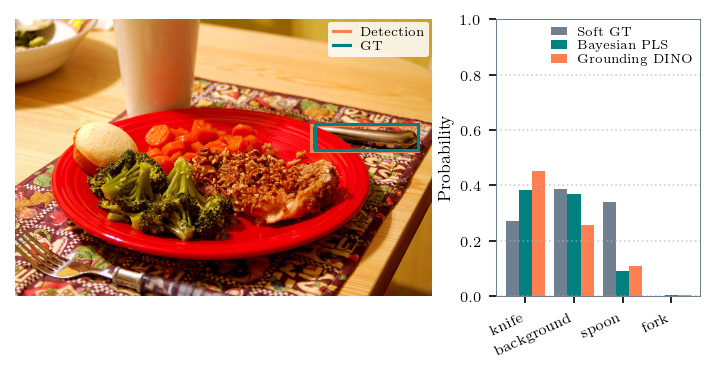}
        \caption{}
        \label{fig:visual-uncertain-gt-3}
    \end{subfigure}
    \caption{Cases where the soft ground truth itself is not fully confident and neither are our method and Grounding DINO.}
    \label{fig:visual-uncertain-gt}
\end{figure*}

\begin{figure*}[p]
    \centering
    \captionsetup[subfigure]{skip=1pt}
    \begin{subfigure}[b]{0.28\textwidth}
        \includegraphics[width=\linewidth]{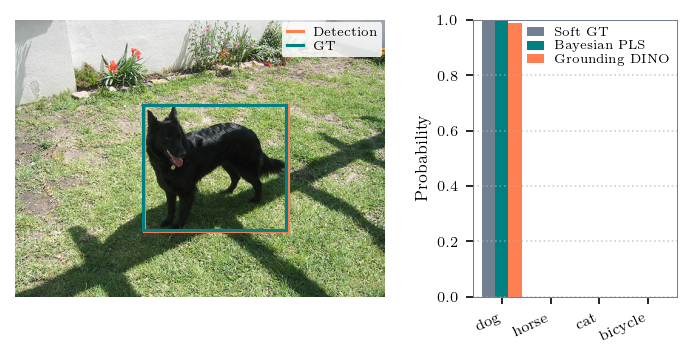}
        \caption{Pascal VOC}
        \label{fig:visual-agreement-voc-0}
    \end{subfigure}
    \hfill
    \begin{subfigure}[b]{0.28\textwidth}
        \includegraphics[width=\linewidth]{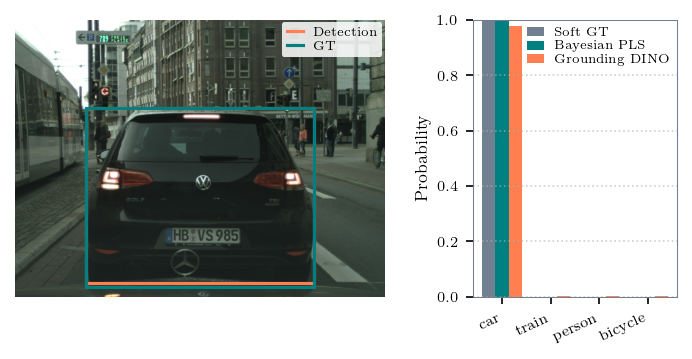}
        \caption{Cityscapes}
        \label{fig:visual-agreement-cityscapes-0}
    \end{subfigure}
    \hfill
    \begin{subfigure}[b]{0.28\textwidth}
        \includegraphics[width=\linewidth]{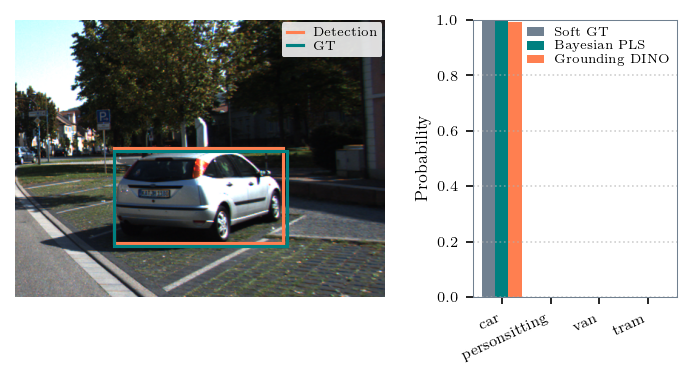}
        \caption{KITTI}
        \label{fig:visual-agreement-kitti-0}
    \end{subfigure}
    \par
    \begin{subfigure}[b]{0.28\textwidth}
        \includegraphics[width=\linewidth]{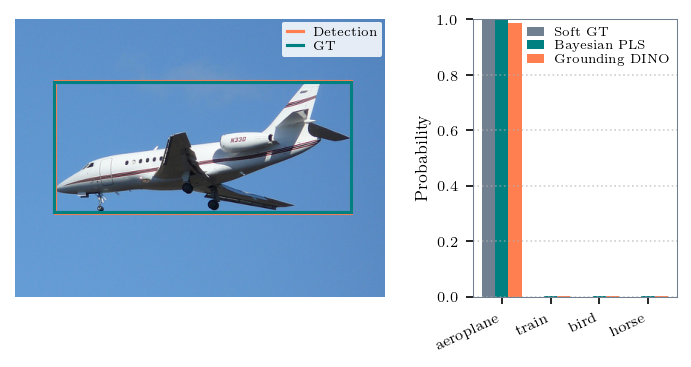}
        \caption{Pascal VOC}
        \label{fig:visual-agreement-voc-1}
    \end{subfigure}
    \hfill
    \begin{subfigure}[b]{0.28\textwidth}
        \includegraphics[width=\linewidth]{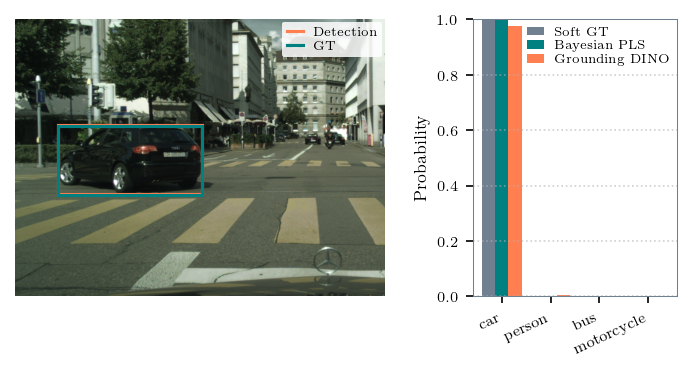}
        \caption{Cityscapes}
        \label{fig:visual-agreement-cityscapes-1}
    \end{subfigure}
    \hfill
    \begin{subfigure}[b]{0.28\textwidth}
        \includegraphics[width=\linewidth]{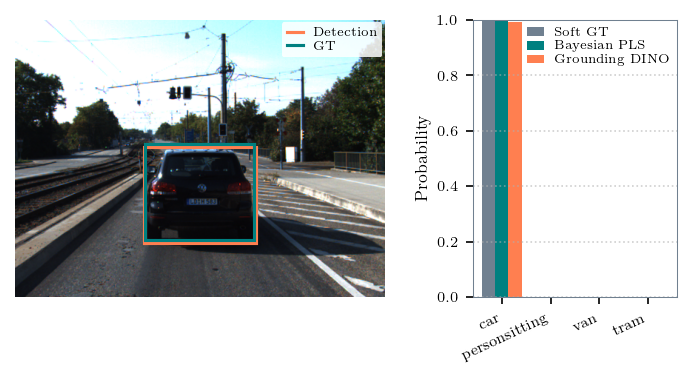}
        \caption{KITTI}
        \label{fig:visual-agreement-kitti-1}
    \end{subfigure}
    \caption{Examples on Pascal VOC, Cityscapes and KITTI where our method and Grounding DINO agree on the prediction.}
    \label{fig:visual-agreement-other}
\end{figure*}

\begin{figure*}[p]
    \centering
    \captionsetup[subfigure]{skip=1pt}
    \begin{subfigure}[b]{0.28\textwidth}
        \includegraphics[width=\linewidth]{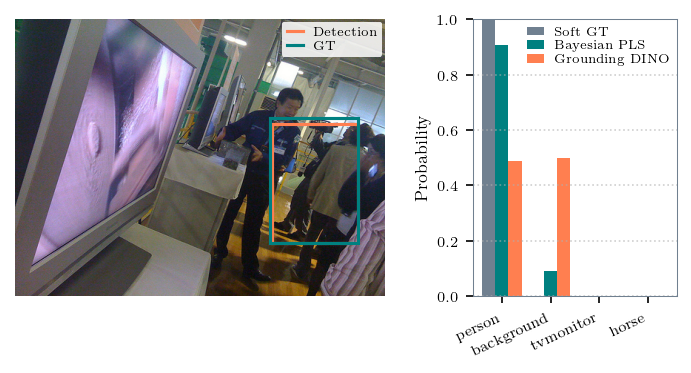}
        \caption{Pascal VOC}
        \label{fig:visual-disagreement-voc-0}
    \end{subfigure}
    \hfill
    \begin{subfigure}[b]{0.28\textwidth}
        \includegraphics[width=\linewidth]{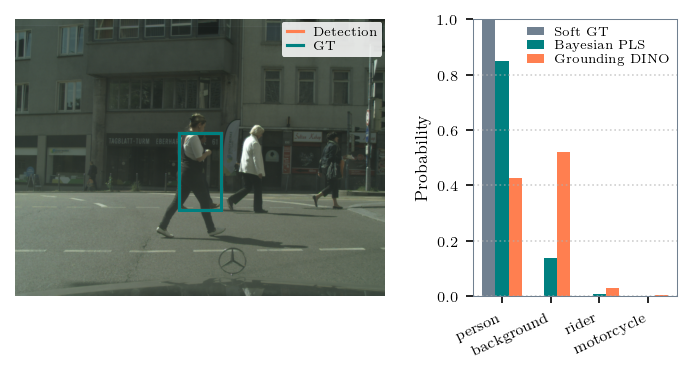}
        \caption{Cityscapes}
        \label{fig:visual-disagreement-cityscapes-0}
    \end{subfigure}
    \hfill
    \begin{subfigure}[b]{0.28\textwidth}
        \includegraphics[width=\linewidth]{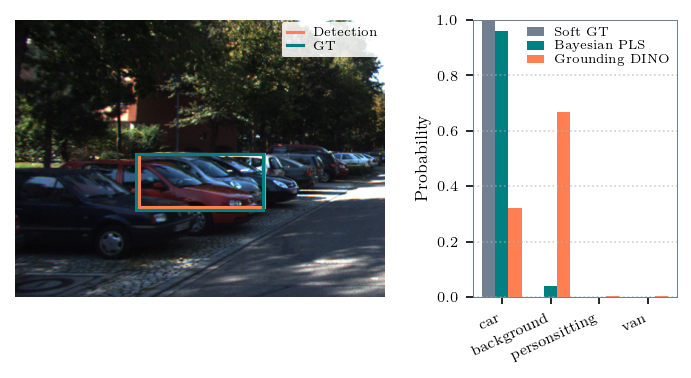}
        \caption{KITTI}
        \label{fig:visual-disagreement-kitti-0}
    \end{subfigure}
    \par
    \begin{subfigure}[b]{0.28\textwidth}
        \includegraphics[width=\linewidth]{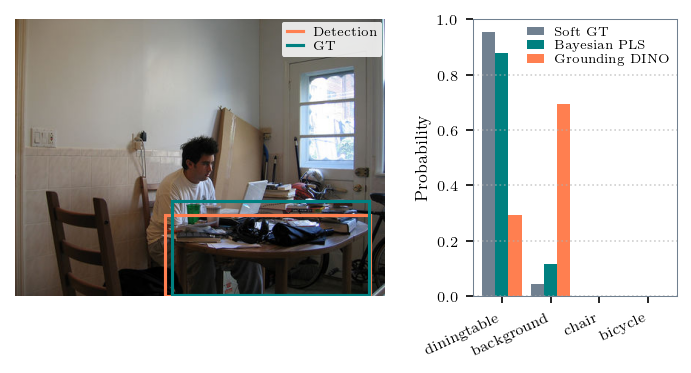}
        \caption{Pascal VOC}
        \label{fig:visual-disagreement-voc-1}
    \end{subfigure}
    \hfill
    \begin{subfigure}[b]{0.28\textwidth}
        \includegraphics[width=\linewidth]{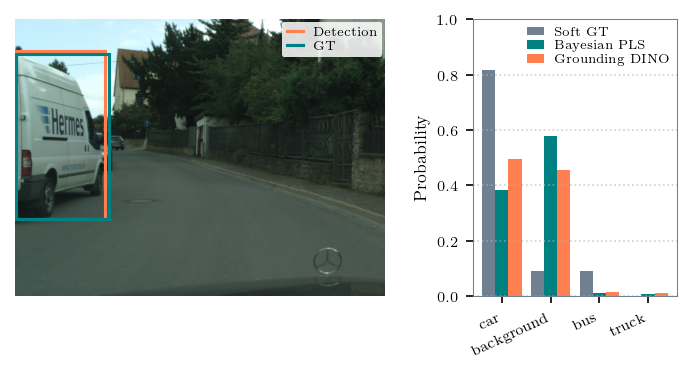}
        \caption{Cityscapes}
        \label{fig:visual-disagreement-cityscapes-1}
    \end{subfigure}
    \hfill
    \begin{subfigure}[b]{0.28\textwidth}
        \includegraphics[width=\linewidth]{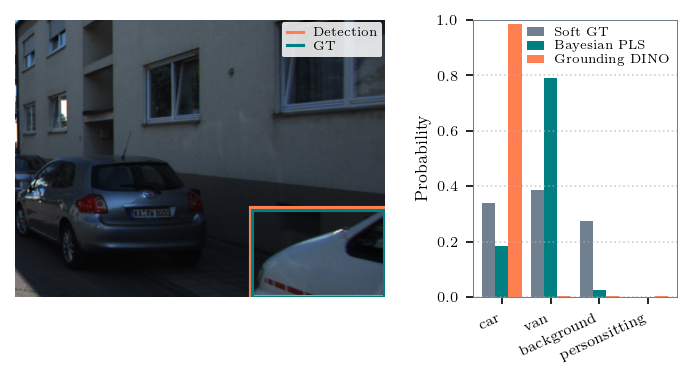}
        \caption{KITTI}
        \label{fig:visual-disagreement-kitti-1}
    \end{subfigure}
    \caption{Examples on Pascal VOC, Cityscapes and KITTI where our method and Grounding DINO do not agree.}
    \label{fig:visual-disagreement-other}
\end{figure*}

\begin{figure*}[p]
    \centering
    \captionsetup[subfigure]{skip=1pt}
    \begin{subfigure}[b]{0.28\textwidth}
        \includegraphics[width=\linewidth]{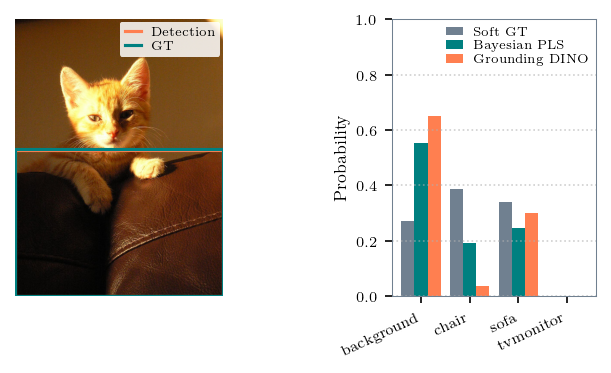}
        \caption{Pascal VOC}
        \label{fig:visual-uncertain-gt-voc-0}
    \end{subfigure}
    \hfill
    \begin{subfigure}[b]{0.28\textwidth}
        \includegraphics[width=\linewidth]{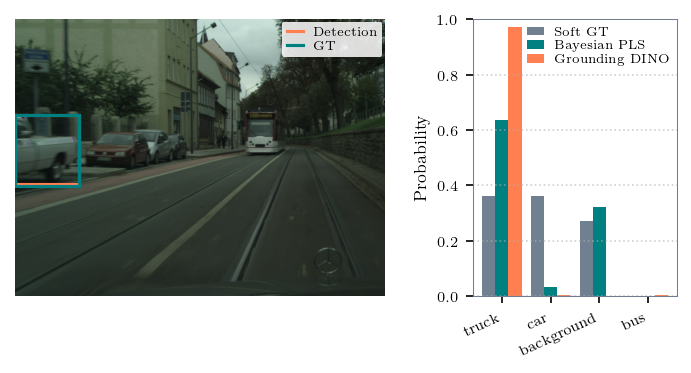}
        \caption{Cityscapes}
        \label{fig:visual-uncertain-gt-cityscapes-0}
    \end{subfigure}
    \hfill
    \begin{subfigure}[b]{0.28\textwidth}
        \includegraphics[width=\linewidth]{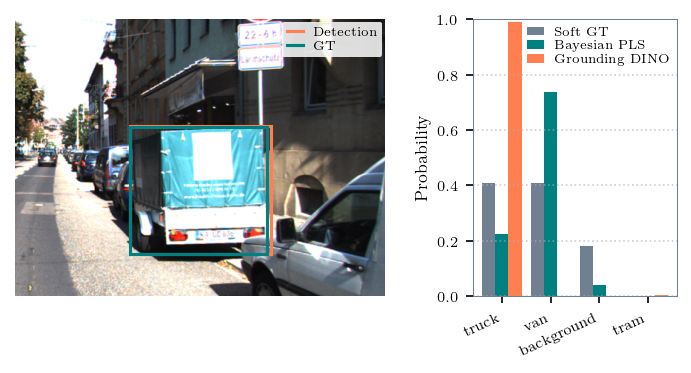}
        \caption{KITTI}
        \label{fig:visual-uncertain-gt-kitti-0}
    \end{subfigure}
    \par
    \begin{subfigure}[b]{0.28\textwidth}
        \includegraphics[width=\linewidth]{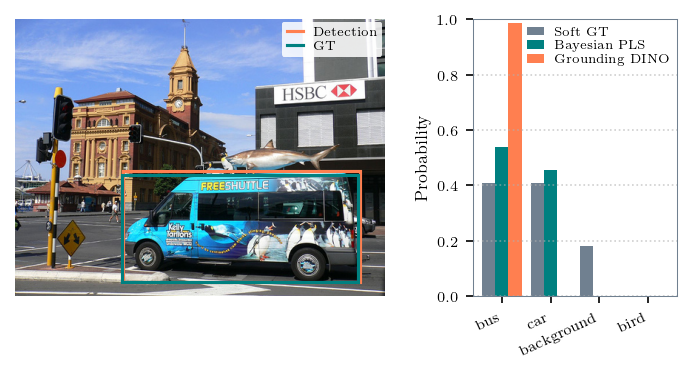}
        \caption{Pascal VOC}
        \label{fig:visual-uncertain-gt-voc-1}
    \end{subfigure}
    \hfill
    \begin{subfigure}[b]{0.28\textwidth}
        \includegraphics[width=\linewidth]{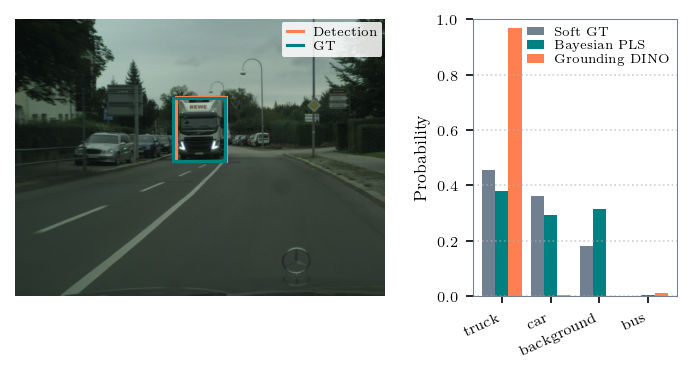}
        \caption{Cityscapes}
        \label{fig:visual-uncertain-gt-cityscapes-1}
    \end{subfigure}
    \hfill
    \begin{subfigure}[b]{0.28\textwidth}
        \includegraphics[width=\linewidth]{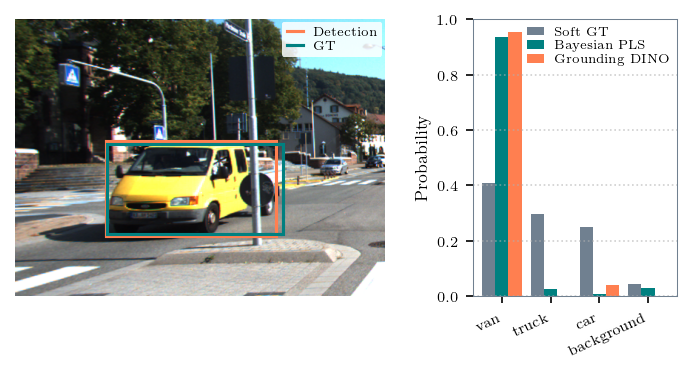}
        \caption{KITTI}
        \label{fig:visual-uncertain-gt-kitti-1}
    \end{subfigure}
    \caption{Relatively uncertain ground truth examples for Pascal VOC, Cityscapes and KITTI.}
    \label{fig:visual-uncertain-gt-other}
\end{figure*}

\end{document}